\documentclass[a4paper,fleqn]{cas-dc}

\usepackage[numbers,sort&compress]{natbib}
\usepackage{booktabs, multirow, tabularx, graphicx, subcaption, stfloats, float, caption}
\newcolumntype{Y}{>{\centering\arraybackslash}X}
\def\tsc#1{\csdef{#1}{\textsc{\lowercase{#1}}\xspace}}
\tsc{WGM}
\tsc{QE}

\begin{document}
\let\WriteBookmarks\relax

\shorttitle{Scale-Robust and Global-Aware Representation Learning for Multivariate Time Series Anomaly Detection}    

\shortauthors{Gyuwon Lee, Saejoon Park, Won-Seok Hwang, Bogoan Kim, and Yunyong Ko}  

\title [mode = title]{Scale-Robust and Global-Aware Representation Learning for Multivariate Time Series Anomaly Detection}  










\affiliation[1]{organization={Chung-Ang University (CAU)},
            city={Seoul},
            country={Republic of Korea}}

\affiliation[2]{organization={The Affiliated Institute of {ETRI}},
            city={Daejeon},
            country={Republic of Korea}}

\affiliation[3]{organization={Chungbuk National University (CBNU)},
            city={Cheongju},
            country={Republic of Korea}}

\author[1]{Gyuwon Lee} [
    orcid=0009-0006-9229-422X
]
\ead{leemarshal@cau.ac.kr}
\credit{Conceptualization, Methodology, Software,
Investigation, Formal analysis, Visualization,
Writing - original draft, Writing - review \& editing, Validation}

\author[1]{Saejoon Park} [
    orcid=0009-0002-9658-7848
]
\ead{kty8300@cau.ac.kr}
\credit{Software, Investigation, Formal analysis}

\author[2]{Won-Seok Hwang} [
    orcid=0009-0001-2925-1239
]
\ead{hws23@nsr.re.kr}
\credit{Conceptualization, Supervision, Writing - review \& editing}

\author[3]{Bogoan Kim} [
    orcid=0000-0002-9083-1128
]
\ead{bogoan@cbnu.ac.kr}
\credit{Supervision, Writing - review \& editing}

\author[1]{Yunyong Ko} [
    orcid=0000-0003-1283-4697
]
\cormark[1]
\ead{yyko@cau.ac.kr}
\credit{Conceptualization, Methodology, Supervision,
Project administration, Writing - review \& editing}

\cortext[1]{Corresponding author}



\newcommand{\method}{\textsf{STAGER}}
\newcommand{\codeurl}{\url{https://github.com/mindslab-cau/STAGER-MTSAD}}
\newcommand{\meanstd}[2]{#1\ensuremath{\pm}{\fontsize{6pt}{7pt}\selectfont #2}}

\begin{abstract}
Multivariate time series anomaly detection (MTSAD) is an essential task for identifying abnormal system behaviors and supporting reliable operations across diverse real-world applications.
Despite substantial advances in MTSAD, two key challenges in representation learning remain underexplored:
\textbf{(C1)} \textit{how to learn scale-robust temporal representations that capture anomalies across different temporal scales} and
\textbf{(C2)} \textit{how to learn global-aware relational representations that capture inter-series dependencies by incorporating series-level global characteristics beyond local windows}.
To address these challenges, we propose {\method}, a novel method that jointly learns two complementary representations: scale-robust temporal and global-aware relational representations. 
Specifically, {\method} employs (1) a scale-robust temporal encoder that constructs a multi-scale temporal directed acyclic graph (DAG) to learn scale-robust temporal representations and (2) a global-aware relational encoder that incorporates learnable series-level global embeddings into the temporal representations and models inter-series relationships through graph neural networks. 
Two complementary auxiliary objectives further enhance the distinctiveness of the global embeddings.
Extensive experiments on three real-world benchmark datasets demonstrate that {\method} achieves the best average rank among ten evaluated MTSAD methods across three evaluation metrics. 
Ablation and sensitivity analyses validate the effectiveness of the proposed components and auxiliary objectives, while efficiency evaluations demonstrate that {\method} achieves a favorable trade-off between detection performance and inference time compared with existing methods.
\end{abstract}





\maketitle

\section{Introduction}\label{sec:intro}
Multivariate time series anomaly detection (MTSAD) aims to identify abnormal observations that deviate from normal patterns in multivariate time series.
As an essential task for intelligent system monitoring and decision-making, it plays a crucial role in understanding system behavior, identifying potential risks, and ensuring operational reliability.
It has been widely applied to diverse real-world domains, including industrial control systems~\cite{mathur2016swat, ahmed2017wadi}, healthcare systems~\cite{moody2001mitbih, schmidt2018wesad}, and financial systems~\cite{li2023deep, POUTRE2024100129, golmohammadi2015time}, to detect system failures, cyber-attacks, and other abnormal events.

Anomalous events are often rare and diverse, making it difficult to collect representative anomaly samples and obtain reliable annotations without extensive domain expertise~\cite{ruszczak2025ops, pang2021deep, zamanzadeh2024deep}.
Consequently, many MTSAD methods adopt an \textit{unsupervised learning} paradigm, in which models learn normal patterns from training data assumed to represent normal system behavior~\cite{hundman2018detecting, deng2021graph}.
During inference, such methods identify unseen observations that deviate from the learned normal patterns as potential anomalies.
Thus, accurate anomaly detection critically depends on how effectively models capture the underlying normal behavior of multivariate time series.

MTSAD methods can be classified into (i) forecasting-based and (ii) reconstruction-based approaches.
Both approaches employ window-based learning, where multivariate time series are divided into fixed-length sliding windows to learn temporal patterns from local observations.
Specifically, (i) forecasting-based methods~\cite{ge2024enhanced, deng2021graph, hundman2018detecting, shen2020timeseries} learn to predict future observations based on preceding windows by minimizing prediction errors, 
whereas (ii) reconstruction-based methods~\cite{shi2023robust, li2025diffusion, yang2026generalizable, audibert2020usad, xu2022anomalytransformer, li2021multivariate, zhang2019deep, xie2025multivariate, chen2025mshtrans} learn to reproduce input windows by minimizing reconstruction errors.
During inference, both approaches compute anomaly scores based on the discrepancies between predicted or reconstructed values and actual observations.
Observations with anomaly scores exceeding a threshold are then detected as anomalies~\cite{deng2021graph, hundman2018detecting, yang2026generalizable}.

\begin{figure*}[tp]
    \centering
    \setlength{\tabcolsep}{1mm}
    \begin{tabular}{@{}cc@{}}
        \includegraphics[width=0.49\linewidth]
        {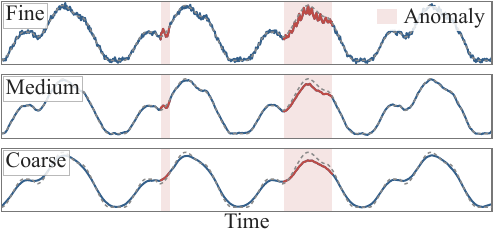} &
        \includegraphics[width=0.49\linewidth]
        {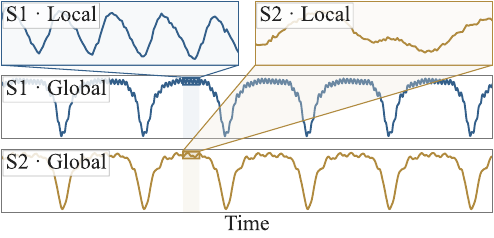} \\
        {\small (a) Challenge 1: Scale-robust temporal representations} &
        {\small (b) Challenge 2: Series-level global representations}
    \end{tabular}    
    \caption{Illustrative examples of two key challenges in MTSAD: (a) scale-robust temporal representations, where short and prolonged anomalies exhibit different visibility across temporal scales, and (b) global-aware relational representations, where two time series exhibit distinct local patterns despite sharing similar long-term periodic behaviors.}
    \label{fig:challenges}
    \vspace{-4mm}
\end{figure*}

\vspace{1mm}
\noindent
\textbf{Challenges.}
Despite breakthroughs in MTSAD, two key challenges in representation learning remain underexplored:
\textbf{(C1)} \textit{how to learn scale-robust temporal representations that capture anomalies across different temporal scales} and
\textbf{(C2)} \textit{how to learn global-aware relational representations that capture inter-series dependencies by incorporating series-level global characteristics beyond local windows}.

\textbf{(C1) Scale-robust temporal representations.} 
Multivariate time series often contain noise and transient fluctuations arising from sensor measurements and dynamic system operations, making it challenging to distinguish meaningful anomalies from normal variations.
Fine-grained representations can preserve subtle and short-lived anomalies but are sensitive to noise, whereas coarse-grained representations suppress noise and capture broader temporal patterns at the risk of missing short anomalies.
As illustrated in Figure~\ref{fig:challenges}(a), a short anomaly becomes less distinguishable at coarser scales due to smoothing, whereas a prolonged anomaly can be more reliably distinguished at coarser scales by suppressing local fluctuations.
Therefore, scale-robust representations should capture temporal patterns while suppressing irrelevant fluctuations and distinguishing anomalies across different temporal scales.

\textbf{(C2) Global-aware relational representations.}
In multivariate time series, individual series often exhibit interdependent behaviors~\cite{li2021multivariate, zhao2020multivariate, zhou2023detecting}, making it essential to capture inter-series relationships for accurate anomaly detection. 
However, existing MTSAD methods typically 
model these relationships primarily using representations from fixed-length windows, which may overlook intrinsic series-level characteristics that persist across windows~\cite{deng2021graph, han2022learning}.
For example, as illustrated in Figure~\ref{fig:challenges}(b), 
two series (S1 and S2) can exhibit different local patterns despite sharing similar long-term periodic behaviors.
Such global characteristics may not be fully captured by window-level representations, 
making it difficult to understand underlying similarities and differences among time series.
Therefore, incorporating series-level global characteristics into relational representations is essential for effectively modeling inter-series dependencies.

\vspace{1mm}
\noindent
\textbf{Our work.}
To address these challenges, in this paper, we propose \textbf{\method} (\textbf{\underline{S}}cale-robust \textbf{\underline{T}}emporal \textbf{\underline{A}}nd \textbf{\underline{G}}lobal-aware r\textbf{\underline{E}}lational \textbf{\underline{R}}epresentation learning), a novel method for multivariate time series anomaly detection.
{\method} employs two encoders, each addressing one challenge: (1) a \textit{scale-robust temporal encoder} that learns temporal representations robust to scale variations (C1), and (2) a \textit{global-aware relational encoder} that models inter-series relationships while incorporating series-level global characteristics beyond local windows (C2).

\textbf{(1) Scale-robust temporal encoder.} 
This module first generates multi-scale time series using moving averages to capture temporal patterns at different scales for each individual series.
It then constructs a temporal directed acyclic graph (DAG)~\cite{thost2021directed} over the multi-scale time series, where each node corresponds to a timestamp and its initial features are formed by concatenating the values of all series across multiple scales at that timestamp.
Finally, a graph neural network (GNN)~\cite{kipf2017semi} is applied to learn scale-robust temporal representations while preserving chronological dependencies.

\textbf{(2) Global-aware relational encoder.} 
Building on the scale-robust temporal representations, 
this module incorporates a learnable series-level global embedding into each series' temporal representations to capture intrinsic characteristics beyond local windows. 
It then constructs an inter-series graph using the globally enhanced representations as node features and applies a graph neural network (GNN) to model inter-series relationships. 
The resulting global-aware relational representations thus integrate scale-robust temporal information, series-level global characteristics, and inter-series relationships. 
Finally, these representations are fed into an anomaly detection decoder to produce predictions for anomaly detection.

Furthermore, to enhance the quality of learnable global embeddings, 
we introduce two complementary auxiliary objectives: 
(a) series classification, which encourages each global embedding to encode series-specific characteristics by predicting its corresponding series identity, 
and (b) distance-based regularization, which prevents embeddings from becoming overly similar by maintaining sufficient separation between their representations.

Our main contributions are summarized as follows.
\begin{itemize}
    \item \textbf{Challenges}: We identify two key challenges in MTSAD: \textbf{(C1)} learning scale-robust temporal representations and \textbf{(C2)} learning global-aware relational representations beyond local windows.
    
    \item \textbf{Method}: We propose {\method} that jointly learns two complementary representations through (1) a scale-robust temporal encoder and (2) a global-aware relational encoder.
    
    \item \textbf{Evaluation}: Extensive experiments on three real-world datasets demonstrate that {\method} achieves the best average rank across three evaluation metrics, while ablation studies validate the effectiveness of each proposed strategy.
\end{itemize}

\begin{table*}[t]
\centering
\caption{Comparison of {\method} with existing methods in terms of the two key challenges (C1) and (C2).}
\label{tab:challenge_comparison}
\setlength{\tabcolsep}{8pt}
\renewcommand{\arraystretch}{1.1}
\renewcommand{\tabularxcolumn}[1]{m{#1}}
\begin{tabularx}{\columnwidth}{@{}>{\raggedright\arraybackslash}Xcc@{}}
\toprule
\multicolumn{1}{c}{\textbf{Methods}}
    & \textbf{(C1)} & \textbf{(C2)} \\
\midrule
kNN, OCSVM, DeepSVDD, LSTM-NDT,
USAD, InterFusion, TSMixer, AnomalyT,
MTAD-GAT, MTGFlow
    & -- & -- \\
\addlinespace[2pt]
MSCRED, MtsCID, MSHTrans
    & $\bigcirc$ & -- \\
\addlinespace[2pt]
GDN, FuSAGNet
    & -- & $\bigcirc$ \\
\midrule
\textbf{\method{} (Ours)}
    & $\bigcirc$ & $\bigcirc$ \\
\bottomrule
\end{tabularx}
\par\smallskip
\end{table*}

\section{Related Works}\label{sec:related}
In this section, we review existing MTSAD methods by categorizing them into three groups: (1) conventional, (2) deep learning-based, and (3) graph-based approaches.
Table~\ref{tab:challenge_comparison} compares {\method} with existing methods in terms of the two key challenges (C1) and (C2).

\subsection{Conventional Methods}
Early anomaly detection methods primarily identify anomalies based on the data distribution or neighborhood structure of normal observations.
kNN~\cite{angiulli2002fast, chen2025pyod} treats the multivariate observation at each time step as a sample and the distance to the $k$-th nearest normal sample as an anomaly score.
OCSVM~\cite{scholkopf2001estimating} estimates a decision boundary enclosing normal data in a kernel space and identifies anomalies based on deviations from the learned boundary.

Although these methods can identify observations that deviate from the normal data distribution, their point-wise formulations fail to fully capture temporal dependencies and inter-series relationships.
Consequently, they have limited ability to characterize normal patterns that depend on temporal evolution and interactions among individual series.

\subsection{Deep Learning-based Methods}
Deep learning-based methods learn representations of normal data, using sliding windows to capture temporal dependencies and complex patterns.
DeepSVDD~\cite{pmlr-v80-ruff18a} learns compact representations of normal data within a hypersphere, where the distance from its center serves as the anomaly score.
LSTM-NDT~\cite{hundman2018detecting} employs independently trained LSTMs~\cite{hochreiter1997long} to forecast each series and applies nonparametric dynamic thresholding to prediction errors.
USAD~\cite{audibert2020usad} learns normal representations through adversarial reconstruction using a shared encoder and two decoders.
InterFusion~\cite{li2021multivariate} employs a hierarchical VAE~\cite{kingma2013auto} to jointly model temporal and inter-series representations.
TSMixer~\cite{chen2023tsmixer} adopts an MLP-Mixer~\cite{tolstikhin2021mlp} architecture to capture temporal and inter-series dependencies.
AnomalyTransformer~\cite{xu2022anomalytransformer} models temporal associations using a Transformer~\cite{vaswani2017attention} architecture and detects anomalies through association discrepancy.
However, these methods do not explicitly account for variations in temporal scale, which may limit their ability to distinguish meaningful anomalies from noise and transient fluctuations across different scales, thus failing to fully address (C1).

To capture temporal dependencies at different temporal scales, MSCRED~\cite{zhang2019deep} constructs multi-scale correlation signature matrices and models their temporal evolution using ConvLSTM~\cite{shi2015convolutional}.
MtsCID~\cite{xie2025multivariate} incorporates frequency-domain information, multi-scale patches, and prototype-based learning to represent normal temporal and relational patterns.
MSHTrans~\cite{chen2025mshtrans} employs trainable multi-scale hypergraphs and time-series decomposition to capture temporal dependencies at different granularities.
These methods, however, learn representations from local windows, which may not sufficiently reflect series-level global characteristics that persist across windows, failing to fully address (C2).

\subsection{Graph-based Methods}
Graph-based MTSAD methods represent individual series as nodes and use graph neural networks (GNNs) to model inter-series dependencies.
MTAD-GAT~\cite{zhao2020multivariate} applies parallel graph attention layers across feature and temporal dimensions and uses a GRU~\cite{cho2014properties} for forecasting and reconstruction.
MTGFlow~\cite{zhou2023detecting} learns a sensor graph for each input window via self-attention and combines this graph with recurrent temporal features to condition an entity-aware normalizing flow.
Both methods infer inter-series relationships primarily from the current window and may therefore overlook similarities in long-term behavior when two series exhibit different local patterns.
GDN and FuSAGNet address (C2) by incorporating sensor-specific embeddings that persist across windows into their estimates of inter-series relationships.
GDN~\cite{deng2021graph} uses these embeddings to select graph neighbors and compute attention weights for forecasting.
FuSAGNet~\cite{han2022learning} selects neighbors based on the cosine similarity of sensor embeddings and uses the embeddings to weight messages derived from sparse autoencoder features.

Furthermore, none of these graph-based methods explicitly accounts for variations in temporal scale, which may limit their ability to learn scale-robust temporal representations (C1).
In contrast, {\method} addresses both challenges by capturing temporal patterns across multiple scales and incorporating series-level characteristics into inter-series relationship modeling.
In Section~\ref{sec:proposed}, we present our proposed method {\method} in detail, explaining how its key components address these challenges.

\begin{figure*}
\centering
\includegraphics[width=1.0\linewidth]{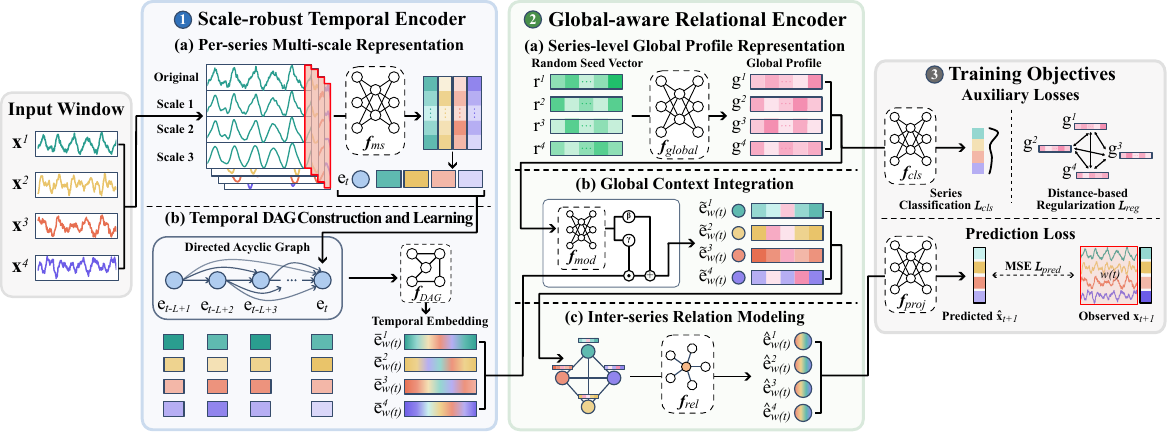}
\caption{Overall architecture of {\method}, consisting of (1) a scale-robust temporal encoder and (2) a global-aware relational encoder. {\method} minimizes a forecasting loss and two auxiliary losses for series classification and distance-based regularization.}
\label{fig:overview}
\end{figure*}

\section{Proposed Method: {\method}}\label{sec:proposed}
In this section, we present {\method} for multivariate time series anomaly detection.
We first introduce the preliminaries and an overview of {\method}.
We then describe the scale-robust temporal encoder and global-aware relational encoder.
Finally, we present the training and inference procedures and analyze the complexity of {\method}.

\subsection{Preliminaries}\label{sec:proposed-prelim}

\vspace{1mm}
\noindent
\textbf{Notations.}
A multivariate time series is denoted by an indexed collection $\{\mathbf{x}^i\}_{i=1}^N$,
where $\mathbf{x}^i$ denotes the $i$-th time series and $N$ is the total number of time series.
Each time series is represented as a sequence of observed values over time, i.e., $\mathbf{x}^i\in\mathbb{R}^{T}$, where $T$ is the number of time steps.
Specifically, $\mathbf{x}^i=[x^i_1,x^i_2,\ldots,x^i_T]$, where $x^i_t$ denotes the observed value of the $i$-th time series at time step $t$.
By stacking the values of all series at time step $t$, the multivariate observation is represented as $\mathbf{x}_t\in\mathbb{R}^N$, where $\mathbf{x}_t=[x^1_t,x^2_t,\ldots,x^N_t]^{\intercal}$.
Accordingly, in matrix form, the multivariate time series is represented as $X=[\mathbf{x}_1,\mathbf{x}_2,\ldots,\mathbf{x}_T]\in\mathbb{R}^{N\times T}$.
MTSAD is typically performed based on a sliding time window.
Given the multivariate observation sequence $[\mathbf{x}_1,\mathbf{x}_2,\ldots,\mathbf{x}_T]$, 
the time window of length $L$ ending at $t$ is denoted by $\mathbf{w}_{(t)} = [\mathbf{x}_{t-L+1}, \mathbf{x}_{t-L+2}, \ldots, \mathbf{x}_{t}]$ for $t\geq L$, where $\mathbf{w}_{(t)}\in\mathbb{R}^{N\times L}$.

The objective of our work is to learn representations of normal patterns and identify anomalous time steps in unseen data.
Accordingly, we formally define the multivariate time series anomaly detection (MTSAD) problem as follows.

\vspace{1mm}
\noindent
\textsc{\textbf{Problem} (Multivariate Time Series Anomaly Detection)}.
\textit{Given a multivariate time series $X\in\mathbb{R}^{N\times T}$, identify the set of anomalous time steps $\mathcal{A}\subseteq\{1,\ldots,T\}$.}

\vspace{1mm}
\noindent
\textbf{Forecasting-based anomaly detection.}
{\method} follows the standard forecasting-based MTSAD approach~\cite{hundman2018detecting,deng2021graph}.
The underlying assumption is that a model trained on normal observations can predict subsequent normal behavior, 
whereas anomalous observations tend to deviate from these predictions.
A forecasting model $\Theta(\cdot)$ predicts the next observation based on the input window: $\hat{\mathbf{x}}_{t+1}=\Theta(\mathbf{w}_{(t)})\in\mathbb{R}^{N}$.
At test time, the anomaly score is computed based on the discrepancy between the predicted values $\hat{\mathbf{x}}_{t+1}$ and the observed values $\mathbf{x}_{t+1}$:
$a_{t+1}=\phi(\hat{\mathbf{x}}_{t+1},\mathbf{x}_{t+1})$, where $\phi(\cdot)$ denotes an error function. 
A larger discrepancy indicates a greater deviation from the learned normal patterns and thus a higher likelihood of anomalous behavior.
An observation is classified as anomalous if $a_{t+1}^{i}>\tau^{i}$, where $\tau^{i}$ is a threshold.

\subsection{Overview}\label{sec:proposed-overview}
As illustrated in Figure~\ref{fig:overview}, {\method} takes a multivariate time window as input and processes it through two encoders:
\begin{itemize}
    \item \textbf{(1) Scale-robust temporal encoder}: constructs a temporal directed acyclic graph (DAG) from multi-scale time series and learns scale-robust temporal representations via a graph neural network (GNN).
    
    \item \textbf{(2) Global-aware relational encoder}: generates learnable series-level global profiles, integrates them into the temporal representations, and learns global-aware relational representations by applying a GNN to an inter-series graph whose initial node features are the globally enhanced representations.
\end{itemize}
The resulting representations are then fed into a decoder to predict the next observation.
During training, {\method} minimizes a forecasting loss along with two auxiliary losses for series classification and distance-based regularization.
During inference, anomalies are identified based on prediction errors between predicted and observed values.

\subsection{Scale-robust Temporal Encoder}\label{sec:proposed-temporal-encoder}
Given an input window $\mathbf{w}_{(t)}\in\mathbb{R}^{N\times L}$,
the scale-robust temporal encoder aims to learn temporal representations that capture diverse patterns at multiple scales while preserving the chronological structure of time series.
To this end, the encoder first generates multi-scale representations for each individual series to capture temporal patterns at different granularities.
It then constructs a temporal directed acyclic graph (DAG) to model chronological dependencies, in which information is propagated only in the forward temporal direction via a graph neural network (GNN).
It consists of two steps:
(a) per-series multi-scale representation and (b) temporal DAG construction and learning.

\vspace{1mm}
\noindent\textbf{{(1)-(a) Per-series multi-scale representation}.}
Original sensor observations may contain measurement errors, transient noise, and operational fluctuations.
Directly modeling such observations can make temporal representations sensitive to local variations, hindering the detection of subtle anomalies.
Applying temporal smoothing may alleviate these disturbances, but the choice of smoothing scale introduces a trade-off between noise suppression and anomaly preservation. 
Specifically, fine-grained representations preserve short-lived anomalies but remain sensitive to noise, whereas coarse-grained representations suppress local fluctuations but may obscure short-lived anomalies as shown in Figure~\ref{fig:challenges}(a). 
Consequently, relying on a single temporal scale may limit the ability to distinguish meaningful anomalies from normal variations.

To address this limitation, we generate multi-scale representations for each individual series by applying moving averages with different window sizes.
This allows the encoder to capture temporal patterns at different granularities, providing complementary information for learning scale-robust temporal representations.

For each series $\mathbf{x}^i\in\mathbb{R}^{L}$ within the input time window $\mathbf{w}_{(t)}$ and a moving-average window length $s$, we generate a smoothed series by applying a moving average:
\begin{equation}
\mathbf{m}_{s}^{i} = \operatorname{MA}_{s}(\mathbf{x}^i) \in\mathbb{R}^{L}.
\end{equation}
Here, $\operatorname{MA}_{s}$ applies a centered moving average within the input window, with endpoint replication padding to preserve the sequence length.
Given a set of scales $\mathcal{S}=\{s_1,\ldots,s_k\}$, the original and smoothed series are stacked to form a multi-scale observation matrix:
\begin{equation}
\mathbf{U}^{i}
=
[\mathbf{x}^{i};
\mathbf{m}_{s_1}^{i};
\ldots;
\mathbf{m}_{s_k}^{i}]
\in\mathbb{R}^{(k+1)\times L}.
\end{equation}
The $t$-th column, $\mathbf{u}_{t}^{i} = [x_t^{i}, m_{t,s_1}^{i}, \ldots, m_{t,s_k}^{i}]^\top\in\mathbb{R}^{k+1}$ contains the multi-scale observations of series $i$ at time step $t$.

To integrate information across different temporal scales, a shared MLP $f_{\mathrm{ms}}$ transforms the multi-scale observations of each series into a $d_m$-dimensional latent representation.
The resulting representations of all $N$ series are concatenated to construct a timestamp-level feature as follows:
\begin{equation}
\mathbf{e}_{t}
=
\mathop{\Vert}_{i=1}^{N}
f_{\mathrm{ms}}(\mathbf{u}_{t}^{i})
\in\mathbb{R}^{Nd_m},
\label{eq:dag_node}
\end{equation}
where $\Vert$ denotes concatenation and $f_{\mathrm{ms}}(\mathbf{u}_{t}^{i})\in\mathbb{R}^{d_m}$ represents the latent representation of series $i$ at time step $t$.
The resulting timestamp-level features incorporate multi-scale information from all series and serve as input features for subsequent chronological dependency learning.


\vspace{1mm}
\noindent\textbf{{(1)-(b) Temporal DAG construction and learning.}}
We then construct a temporal directed acyclic graph (DAG) to model chronological dependencies within the input window.
In this temporal DAG, each time step corresponds to a node with feature $\mathbf{e}_t\in\mathbb{R}^{Nd_m}$, 
and directed edges allow information to propagate only from earlier to later time steps, thereby preserving temporal chronology.

Let $\mathbf{A}\in\mathbb{R}^{L\times L}$ denote the adjacency matrix of the temporal DAG.
Each element $A_{ij}$ represents the edge weight from source node $j$ to destination node $i$, defined as 
\begin{equation}
A_{ij}
=
\begin{cases}
\sigma(\delta+j-i), & j\leq i,\\
0, & j>i,
\end{cases}
\label{eq:dag_adj}
\end{equation}
where $\sigma(\cdot)$ is the sigmoid function and $\delta$ controls the effective temporal range of historical influence.
This lower-triangular structure allows each timestamp to aggregate information only from itself and preceding timestamps.
We row-normalize the adjacency matrix as $\widetilde{\mathbf{A}} = \mathbf{D}^{-1}\mathbf{A}$, where  $D_{ii}=\sum_j A_{ij}$.

We then stack the timestamp-level features in chronological order to form the initial node feature matrix $\mathbf{E}^{(0)}=[\mathbf{e}_{t-L+1},\ldots,\mathbf{e}_{t}]^\top\in\mathbb{R}^{L\times Nd_m}$.
An $M$-layer GCN~\cite{kipf2017semi} is applied to the temporal DAG to capture the chronological dependencies of multivariate time series.
At each layer, the node representations are updated by aggregating information from the current and preceding timestamps according to the normalized adjacency matrix as
\begin{equation}
\mathbf{E}^{(\ell+1)}
=
\rho\!\left(
\widetilde{\mathbf{A}}
\mathbf{E}^{(\ell)}
\mathbf{W}^{(\ell)}
\right),
\label{eq:dag_prop}
\end{equation}
where $\mathbf{W}^{(\ell)}$ is the learnable weight matrix of the $\ell$-th layer and $\rho(\cdot)$ denotes a nonlinear activation function (ReLU).
The resulting temporal representations from the $M$-layer GCN are denoted by 
\begin{equation}
\mathbf{E}^{(M)}
=
f_{\mathrm{DAG}}(\mathbf{E}^{(0)};\widetilde{\mathbf{A}})
\in\mathbb{R}^{L\times Nd_t}.
\label{eq:dag_encoder}
\end{equation}

Finally, we aggregate the timestamp-level representations by mean pooling over the temporal dimension:
\begin{equation}
\bar{\mathbf{e}}_{\mathbf{w}_{(t)}}
=
\frac{1}{L}
\sum_{r=1}^{L}
\mathbf{E}_{r}^{(M)}
\in\mathbb{R}^{Nd_t}.
\label{eq:temporal_pool}
\end{equation}
The pooled representation summarizes the temporal information of the entire window while preserving the concatenated representations of all series.
It is then reshaped into the series-wise temporal representations:
\begin{equation}
\bar{\mathbf{E}}_{\mathbf{w}_{(t)}}
=
[
\bar{\mathbf{e}}_{\mathbf{w}_{(t)}}^{\,1};
\ldots;
\bar{\mathbf{e}}_{\mathbf{w}_{(t)}}^{\,N}
]
\in\mathbb{R}^{N\times d_t},
\label{eq:temporal_rep}
\end{equation}
where $\bar{\mathbf{e}}_{\mathbf{w}_{(t)}}^{\,i}\in\mathbb{R}^{d_t}$ corresponds to the representation of the $i$-th series that captures multi-scale temporal patterns and chronological dependencies.

Through multi-scale representation learning and temporal DAG-based aggregation, the resulting representations integrate temporal patterns at different granularities with information from preceding time steps, providing a comprehensive characterization of the temporal behavior within each input window.
These representations are further enriched by the global-aware relational encoder, as described in Section~\ref{sec:proposed-relational-encoder}.

\subsection{Global-aware Relational Encoder}\label{sec:proposed-relational-encoder}
The global-aware relational encoder aims to learn inter-series relational representations by incorporating global characteristics of individual series that may not be fully captured within a limited input window.
Specifically, this encoder first learns a global profile for each series to represent its persistent characteristics beyond local temporal observations. 
It then integrates these global profiles with the temporal representations $\bar{\mathbf{E}}_{\mathbf{w}_{(t)}}$ obtained from the scale-robust temporal encoder. 
Finally, a second GNN is applied to an inter-series graph to capture dependencies among different series based on the globally enhanced representations.
It consists of three steps:
(a) series-level global profile representation, (b) global context integration, and (c) inter-series relational modeling.

\vspace{1mm}
\noindent\textbf{{(2)-(a) Series-level global profile representation.}}
Since global characteristics may not be fully captured within limited input windows and no predefined global descriptor for each series is generally available, 
we learn series-specific global embeddings to represent the global profile of each series.
Let $\mathbf{r}^{i}\in\mathbb{R}^{d_s}$ denote a random seed vector for the $i$-th series, initialized once and kept fixed throughout training and inference.
A shared MLP transforms the seed vector into a global embedding:
\begin{equation}
\mathbf{g}^{i}
=
f_{\mathrm{global}}(\mathbf{r}^{i})
\in\mathbb{R}^{d_g}.
\end{equation}
Although the seed vectors are fixed, the parameters of $f_{\mathrm{global}}$ are learned to produce global embeddings that capture series-specific characteristics.
The same global embedding $\mathbf{g}^{i}$ is shared across all windows of the $i$-th series, allowing it to receive training signals from multiple windows and learn series-specific characteristics useful for forecasting.
It thus serves as a window-independent series-level representation, complementing the temporal representation of each input window.
By stacking all series embeddings, we obtain 
\begin{equation}
\mathbf{G}
=
[
\mathbf{g}^{1};
\mathbf{g}^{2};
\ldots;
\mathbf{g}^{N}
]
\in\mathbb{R}^{N\times d_g}.
\end{equation}
To encourage the global embeddings to capture discriminative characteristics of individual series, we introduce two auxiliary objectives: 
(i) \textit{series classification} and (ii) \textit{distance-based regularization}, as described in Section~\ref{sec:proposed-training}.

\vspace{1mm}
\noindent\textbf{(2)-(b) Global context integration.}
For each series, we integrate the temporal representation from Section~\ref{sec:proposed-temporal-encoder} with its corresponding global embedding, producing a representation that encodes both local temporal patterns and series-level global characteristics.
The integrated representation is computed as 
\begin{equation}
\tilde{\mathbf{e}}_{\mathbf{w}_{(t)}}^{\,i}
=
f_{\mathrm{fuse}}
\left(
\bar{\mathbf{e}}_{\mathbf{w}_{(t)}}^{\,i},
\mathbf{g}^{i}
\right)
\in\mathbb{R}^{d_f},
\label{eq:global_fusion}
\end{equation}
where $f_{\mathrm{fuse}}(\cdot,\cdot)$ denotes a fusion function that integrates the temporal representation $\bar{\mathbf{e}}_{\mathbf{w}_{(t)}}^{\,i}$ with the corresponding global embedding $\mathbf{g}^{i}$.
In practice, $f_{\mathrm{fuse}}$ can be implemented using various fusion mechanisms, such as feature concatenation followed by an MLP.

In this work, we adopt Feature-wise Linear Modulation (FiLM)~\cite{perez2018film} as the fusion mechanism,
which generates feature-wise scaling and bias vectors from the global embedding to modulate the temporal representation as follows:
\begin{equation}
f_{\mathrm{fuse}}
\left(
\bar{\mathbf{e}}_{\mathbf{w}_{(t)}}^{\,i},
\mathbf{g}^{i}
\right)
=
\gamma(\mathbf{g}^{i})
\odot
\bar{\mathbf{e}}_{\mathbf{w}_{(t)}}^{\,i}
+
\beta(\mathbf{g}^{i}),
\end{equation}
where $\gamma,\beta: \mathbb{R}^{d_g}\rightarrow\mathbb{R}^{d_t}$ (with $d_f=d_t$) are learnable mappings that transform the global embedding into feature-wise scaling and bias vectors, respectively. 
$\odot$ denotes element-wise multiplication.

Applying the fusion to all series yields the global context-enhanced representations
\begin{equation}
\widetilde{\mathbf{E}}_{\mathbf{w}_{(t)}}
=
[
\tilde{\mathbf{e}}_{\mathbf{w}_{(t)}}^{\,1};
\ldots;
\tilde{\mathbf{e}}_{\mathbf{w}_{(t)}}^{\,N}
]
\in\mathbb{R}^{N\times d_f}.
\end{equation}
Through this modulation, the temporal representation incorporates series-level global characteristics beyond the input window.
The subsequent relational modeling therefore operates on representations conditioned on both the current temporal patterns and the series-level global context.

\noindent
\vspace{2mm}
\textbf{(2)-(c) Inter-series relational modeling.}
Given the global context-enhanced representations
$\widetilde{\mathbf{E}}_{\mathbf{w}_{(t)}}\in\mathbb{R}^{N\times d_f}$,
the relational encoder captures inter-series dependencies among series.
We construct a complete inter-series graph where each node corresponds to a series and its initial node feature is the corresponding global context-enhanced representation.

To learn inter-series dependencies on this graph, we apply a Graph Attention Network (GAT)~\cite{velivckovic2017graph},
which assigns different attention weights to neighboring series based on their node representations.
Unlike uniform aggregation, the attention mechanism allows the encoder to capture varying strengths of dependencies among series.
The resulting relational representations are computed as
\begin{equation}
\hat{\mathbf{E}}_{\mathbf{w}_{(t)}}
=
f_{\mathrm{rel}}
\!\left(
\widetilde{\mathbf{E}}_{\mathbf{w}_{(t)}}
\right)
\in\mathbb{R}^{N\times d_r},
\label{eq:relation_encoder}
\end{equation}
where $f_{\mathrm{rel}}(\cdot)$ denotes a GAT-based relational encoder.

By integrating series-level global characteristics with window-specific temporal patterns and modeling their inter-series dependencies, the global-aware relational encoder produces representations that capture temporal, global, and relational information in a unified manner. 
These representations serve as the final output of the proposed representation learning framework and are subsequently used for forecasting-based training and anomaly detection during inference, as described in Section~\ref{sec:proposed-training}.

\subsection{Training and Inference}\label{sec:proposed-training}
The final representations $\hat{\mathbf{E}}_{(t)}$ are fed into an MLP-based prediction head to forecast the next observation:
\begin{equation}
\hat{\mathbf{x}}_{t+1}
=
f_{\mathrm{proj}}
\!\left(
\hat{\mathbf{E}}_{\mathbf{w}_{(t)}}
\right)
\in\mathbb{R}^{N},
\label{eq:prediction}
\end{equation}
where $f_{\mathrm{proj}}(\cdot)$ is a single-layer MLP.
The parameters of {\method} are primarily trained to minimize the forecasting error between the predicted and ground-truth observations.
The primary forecasting loss is defined using the mean squared error (MSE):
\begin{equation}
L_{\mathrm{pred}}
=
\frac{1}{N}
\sum_{i=1}^{N}
\left(
\hat{x}_{t+1}^{\,i}
-
x_{t+1}^{\,i}
\right)^2.
\label{eq:pred_loss}
\end{equation}

In addition to the forecasting loss, we introduce two auxiliary losses to learn the global embeddings that capture discriminative series-level characteristics: (i) series classification and (ii) distance-based regularization.
The series classification task encourages each global embedding to retain information from which its corresponding series identity can be predicted.
Specifically, given the global embedding $\mathbf{g}^{\,i}$, an auxiliary MLP classifier $f_{\mathrm{cls}}(\cdot)$ predicts the identity of its corresponding series.
The classification loss is defined as the cross-entropy loss:
\begin{equation}
L_{\mathrm{cls}}
=
\frac{1}{N}
\sum_{i=1}^{N}
\mathrm{CE}
\!\left(
f_{\mathrm{cls}}(\mathbf g^{\,i}),
i
\right).
\end{equation}

Series classification encourages global embeddings to encode series identity, while distance-based regularization promotes their separation directly in the embedding space.
The regularization term is defined using the average pairwise distance between normalized global embeddings:
\begin{equation}
L_{\mathrm{reg}} = 2- 
\frac{2}{N(N-1)}
\sum_{1\le i<j\le N}
\left\|
\bar{\mathbf g}^{\,i}
-
\bar{\mathbf g}^{\,j}
\right\|_2,
\end{equation}
where
$\bar{\mathbf g}^{\,i}
=
\mathbf g^{\,i}/
\|\mathbf g^{\,i}\|_2$.
Since the embeddings are normalized to unit length, their pairwise Euclidean distances are bounded above by $2$.
Minimizing $L_{\mathrm{reg}}$ therefore increases the average pairwise distance between global embeddings, promoting their separation in the embedding space.
The total training loss is defined as follows:
\begin{equation}
L
=
L_{\mathrm{pred}}
+
\alpha L_{\mathrm{reg}}
+
\beta L_{\mathrm{cls}},
\end{equation}
where $\alpha$ and $\beta$ control the contributions of the two auxiliary losses. 
We evaluate the impact of the auxiliary losses on the performance of {\method} under different hyperparameter settings in Section~\ref{sec:eq3}.

During inference, the trained model predicts the next observation for each input window and computes anomaly scores based on the prediction errors.
For the $i$-th series, the anomaly score at time step $t+1$ is defined as the squared prediction error:
\begin{equation}
a_{t+1}^{\,i}
=
\left(
\hat{x}_{t+1}^{\,i}
-
x_{t+1}^{\,i}
\right)^2.
\label{eq:series_anomaly_score}
\end{equation}
Each series is independently classified as normal or anomalous by comparing its anomaly score with a series-specific threshold $\tau^{i}$:
\begin{equation}
\hat{y}_{t+1}^{\,i}
=
\begin{cases}
1, & a_{t+1}^{\,i}>\tau^{i},\\
0, & \text{otherwise},
\end{cases}
\end{equation}
where $\hat{y}_{t+1}^{\,i}=1$ indicates that the $i$-th series is detected as anomalous.
Finally, a timestamp is classified as anomalous if at least one series is detected as anomalous:
\begin{equation}
\hat{y}_{t+1}
=
\bigvee_{i=1}^{N}
\hat{y}_{t+1}^{\,i},
\end{equation}
where $\vee$ denotes the logical OR operator.

\subsection{Computational Complexity}\label{sec:proposed-complexity}
In this section, we analyze the computational cost of {\method} for processing each input window.
Let $L$, $N$, and $d$ denote the window length, the number of series, and the embedding dimension, respectively. 
For simplicity, we assume a common embedding dimension $d$ across all components, with hidden widths proportional to $d$. 
We also treat the number of moving-average scales, temporal GCN layers, MLP layers, and attention heads as constants.
The multi-scale representation requires $\mathcal{O}(LNd^2)$ operations.
Each temporal GCN layer performs temporal aggregation in $\mathcal{O}(L^2Nd)$ time and a dense transformation of the concatenated series representations in $\mathcal{O}(LN^2d^2)$ time.
The global context encoder and the relational GAT on a complete graph require $\mathcal{O}(Nd^2)$ and $\mathcal{O}((Nd^2+N^2d))$ operations, respectively.
During training, the auxiliary objectives introduce an additional computational cost of $\mathcal{O}((Nd^2+N^2d))$ per window, without changing the leading order.
Since backpropagation preserves the asymptotic order of the forward computation,
the overall per-window time complexity is
\begin{equation}
\mathcal{O}\!\left(L^2Nd+LN^2d^2\right).
\end{equation}

Therefore, the overall complexity of {\method} is primarily determined by the window length $L$ and the number of series $N$, exhibiting quadratic scaling with respect to both variables.
This scaling behavior is comparable to that of existing MTSAD methods~\cite{han2022learning, chen2025mshtrans}.
Table~\ref{tab:complexity} summarizes the per-window time complexity of {\method} and existing MTSAD methods for comparison.
While theoretical complexity provides an asymptotic characterization of computational cost, actual efficiency can be influenced by implementation and hardware factors. 
To complement this theoretical analysis, 
we empirically compare the computational efficiency of {\method} with existing methods in Section~\ref{sec:eq4}.


\begin{table}[htbp]
\caption{Per-window complexity of {\method} and existing methods.}
\label{tab:complexity}
\centering
\setlength{\tabcolsep}{10pt}
\begin{tabular}{@{}lc@{}}
\toprule
Method & Time Complexity \\
\midrule
\method{} & $\mathcal{O}\!\left(L^2Nd+LN^2d^2\right)$ \\
USAD & $\mathcal{O}\!\left(L^2N^2+LNd\right)$ \\
TSMixer & $\mathcal{O}\!\left(L^2N+LNd\right)$ \\
Anomaly Transformer & $\mathcal{O}\!\left(L^2d+LNd+Ld^2\right)$ \\
FuSAGNet & $\mathcal{O}\!\left(L^2N^2+LN^2d+N^2d^2\right)$ \\
MtsCID & $\mathcal{O}\!\left(L^2d+LN^2+Ld^2\right)$ \\
MSHTrans & $\mathcal{O}\!\left(L^2N+LN^2+LNd\right)$ \\
\bottomrule
\end{tabular}
\end{table}


\section{Experimental Validation}\label{sec:eval}
In this section, we comprehensively evaluate {\method} by answering the following evaluation questions (EQs):
\begin{itemize}
    \item \textbf{EQ1.} To what extent does {\method} improve anomaly detection performance over existing methods?
    \item \textbf{EQ2.} How effective is each of our proposed components in improving detection performance?
    \item \textbf{EQ3.} How do our auxiliary objectives contribute to the detection performance of {\method}?
    \item \textbf{EQ4.} How efficient is {\method} in terms of the trade-off between detection performance and inference time?
\end{itemize}

\subsection{Experimental Setup}

\begin{table}[t]
\caption{Statistics of the datasets used in our experiments.}
\label{tab:dataset_stats}
\centering
\setlength{\tabcolsep}{8pt}
\begin{tabular}{ccccc}
\toprule
Dataset & \#Series & \#Train & \#Test & Anomaly \\
\midrule
SWaT & 51 & 47,340 & 44,993 & 12.14\% \\
WADI & 127 & 76,299 & 17,281 & 5.77\% \\
HAI & 86 & 100,440 & 36,120 & 3.32\% \\
\bottomrule
\end{tabular}
\end{table}

\noindent\textbf{{Datasets.}}
We evaluate {\method} on three real-world industrial control system (ICS) datasets: 
Secure Water Treatment (SWaT)~\cite{mathur2016swat} and Water Distribution (WADI)~\cite{ahmed2017wadi},
and HIL-based Augmented ICS (HAI)~\cite{shin2023hai}.
Specifically, we use SWaT.A2, WADI.A2, and HAI 22.04, respectively.
These datasets contain multivariate sensor measurements and cyber-attack scenarios from different ICS environments.
Following prior studies~\cite{deng2021graph,han2022learning}, 
we downsample all datasets to a 10-second resolution, using the median and mode to aggregate sensor values and anomaly labels, respectively.
We additionally remove the first 2,160 samples (about six hours) of training data from SWaT and WADI to exclude unstable system initialization.
Table~\ref{tab:dataset_stats} summarizes the dataset statistics after preprocessing.


\vspace{1mm}
\noindent\textbf{{Competing methods.}}
We compare {\method} with nine existing multivariate time series anomaly detection methods:
kNN~\cite{angiulli2002fast}, OCSVM \cite{scholkopf2001estimating}, DeepSVDD~\cite{pmlr-v80-ruff18a}, USAD~\cite{audibert2020usad}, TSMixer~\cite{chen2023tsmixer}, AnomalyTransformer~\cite{xu2022anomalytransformer}, FuSAGNet~\cite{han2022learning}, MtsCID~\cite{xie2025multivariate}, and MSHTrans~\cite{chen2025mshtrans}.
For fair comparison, we use the official implementations provided by the original authors whenever available, 
with kNN implemented using the PyOD library~\cite{chen2025pyod}. 

\vspace{1mm}
\noindent\textbf{{Evaluation protocol.}}
For each model, anomaly scores are computed for individual time steps in the test set, and anomalies are identified by applying a threshold to these scores.  
Following~\cite{hwang2026n41}, we select the threshold that maximizes the eTaF1 score with a false positive rate (FPR) constraint of 0.002. 
To comprehensively evaluate anomaly detection performance,
we employ three complementary metrics: (1) conventional precision(P)/recall(R)/F1, (2) point-adjusted (PA) P/R/F1~\cite{xu2018unsupervised}, and (3) enhanced time-series aware (eTa) P/R/F1~\cite{hwang2022you}. 
We report the average results over five runs with different random seeds.

\vspace{1mm}
\noindent\textbf{{System Configuration.}}
All experiments are conducted on Ubuntu 22.04.4 LTS with Python 3.11.0.
Deep learning-based methods are implemented using PyTorch 2.4.0, PyTorch Geometric 2.6.1, and CUDA 11.8, 
while classical baselines are implemented using PyOD 2.0.5.
Experiments are performed on a machine equipped with an AMD Ryzen 9 9900X CPU, two NVIDIA GeForce RTX 4080 SUPER GPUs (16~GB each), and 64~GB RAM.

\begin{table}[t]
\caption{Dataset-specific hyperparameter configurations of {\method}.}
\label{tab:hyperparameters}
\centering
\setlength{\tabcolsep}{1pt}
\def\arraystretch{0.9}
\begin{tabularx}{\columnwidth}{
@{}
>{\hsize=1.6\hsize\linewidth=\hsize
  \raggedright\arraybackslash}X
>{\hsize=0.8\hsize\linewidth=\hsize
  \centering\arraybackslash}X
>{\hsize=0.8\hsize\linewidth=\hsize
  \centering\arraybackslash}X
>{\hsize=0.8\hsize\linewidth=\hsize
  \centering\arraybackslash}X
@{}
}
\toprule
Hyperparameter & SWaT & WADI & HAI \\
\midrule
Window length \(L\)
& 120 & 360 & 15 \\

Embedding dimension \(d\)
& 16 & 16 & 16 \\

Batch size
& 32 & 32 & 32 \\

Learning rate
& \(10^{-4}\) & \(10^{-4}\) & \(10^{-4}\) \\

Training epochs (max.)
& 50 & 50 & 50 \\

\(f_{\mathrm{DAG}}\)\(M\)-layers
& 2 & 2 & 2 \\

\(f_{\mathrm{rel}}\) layers
& 1 & 1 & 1 \\

\(f_{\mathrm{rel}}\) heads
& 4 & 1 & 4 \\

\(f_{\mathrm{rel}}\) dropout
& 0.1 & 0.1 & 0.1 \\

Multi-scale windows \(\mathcal{S}\)
& \([5,10,20]\)
& \([5,10,20]\)
& \([5,10,15]\) \\

Aux. weights \((\alpha,\beta)\)
& \((0.3,0.1)\)
& \((0.2,0.3)\)
& \((0.2,0.2)\) \\
\bottomrule
\end{tabularx}
\end{table}

\begin{table}[!t]
\caption{Input window sizes used by {\method} and the baseline methods.}
\label{tab:window_sizes}
\centering
\setlength{\tabcolsep}{11.5pt}
\def\arraystretch{0.9} 
\begin{tabular}{lccc}
\toprule
Method & SWaT & WADI & HAI \\
\midrule
\method{}          & 120 & 360 & 15 \\
USAD               &  12 &  10 & 15 \\
TSMixer            &  60 & 300 & 15 \\
AnomalyTransformer & 100 &  60 & 15 \\
FuSAGNet           &  16 &  60 & 16 \\
MtsCID             & 100 &  30 & 15 \\
MSHTrans           & 100 & 100 & 15 \\
\bottomrule
\end{tabular}
\end{table}

\vspace{1mm}
\noindent\textbf{Implementation details.}
We use the AdamW~\cite{DBLP:conf/iclr/LoshchilovH19} optimizer with a learning rate of $10^{-4}$, a batch size of 32, and early stopping with a patience of 10 epochs (maximum 50 epochs).
Across all three datasets, we use a 16-dimensional embedding, two temporal GCN layers, and one relational GAT layer.
The input window length $L$, multi-scale moving-average windows $\mathcal{S}$, number of GAT attention heads, and auxiliary-objective weights $(\alpha,\beta)$ are configured separately for each dataset.
Table~\ref{tab:hyperparameters} summarizes the dataset-specific hyperparameter configurations of {\method}.
For the baselines, we adopt the original hyperparameter settings whenever available and otherwise use tuned configurations.
In particular, we use the input window lengths reported in the original works or select better-performing values through additional tuning.
Table~\ref{tab:window_sizes} summarizes the input window lengths used by {\method} and the baseline methods.

For reproducibility, we provide our source code and detailed information on the datasets, baseline methods, and implementation settings, including hyperparameter configurations, at {\codeurl}.

\begin{table*}[!t]
\caption{Anomaly detection performance on SWaT, WADI, and HAI datasets. {\method} achieves the best average ranking across F1, PA-F1, and eTaF1 on the three datasets. The \textbf{bold} and \underline{underline} indicate the best and second-best results, respectively.}\label{tbl:full_results}
\centering
\footnotesize
\setlength{\tabcolsep}{1pt}
\def\arraystretch{1.05} 
\begin{tabularx}{\textwidth}{clYYYYYYYYYY}
\toprule
Dataset & Metric & KNN & OCSVM & DeepSVDD & USAD & TSMixer & AnomalyT & FuSAGNet & MtsCID & MSHTrans & \method{} \\
\midrule
\multirow[c]{10}{*}{SWaT}
& P    & \meanstd{97.89}{0.00} & \meanstd{97.88}{0.00} & \meanstd{97.64}{0.03} & \meanstd{98.12}{0.08} & \meanstd{97.89}{0.03} & \meanstd{97.89}{0.08} & \underline{\meanstd{98.17}{0.05}} & \meanstd{34.36}{0.75} & \meanstd{98.09}{0.02} & \textbf{\meanstd{98.24}{0.04}} \\
& R       & \meanstd{65.26}{0.00} & \meanstd{64.96}{0.00} & \meanstd{58.29}{0.16} & \meanstd{71.08}{1.51} & \meanstd{65.34}{1.01} & \meanstd{65.51}{2.63} & \underline{\meanstd{75.64}{1.99}} & \meanstd{0.73}{0.02} & \meanstd{71.54}{0.84} & \textbf{\meanstd{78.55}{1.61}} \\
& F1           & \meanstd{78.31}{0.00} & \meanstd{78.09}{0.00} & \meanstd{73.00}{0.13} & \meanstd{82.43}{1.00} & \meanstd{78.36}{0.74} & \meanstd{78.47}{1.92} & \underline{\meanstd{85.44}{1.29}} & \meanstd{1.43}{0.05} & \meanstd{82.73}{0.57} & \textbf{\meanstd{87.29}{1.01}} \\
\cmidrule(lr){2-12}
& PA-P & \meanstd{98.14}{0.00} & \meanstd{98.14}{0.00} & \meanstd{97.95}{0.02} & \meanstd{98.37}{0.06} & \textbf{\meanstd{98.47}{0.01}} & \meanstd{98.22}{0.08} & \underline{\meanstd{98.44}{0.02}} & \meanstd{98.14}{0.03} & \meanstd{98.38}{0.03} & \textbf{\meanstd{98.47}{0.03}} \\
& PA-R    & \meanstd{74.35}{0.00} & \meanstd{74.35}{0.00} & \meanstd{67.06}{0.42} & \meanstd{81.94}{2.87} & \textbf{\meanstd{90.85}{0.57}} & \meanstd{78.12}{3.52} & \meanstd{88.95}{1.12} & \meanstd{73.75}{0.68} & \meanstd{84.58}{1.79} & \underline{\meanstd{90.50}{2.06}} \\
& PA-F1        & \meanstd{84.61}{0.00} & \meanstd{84.61}{0.00} & \meanstd{79.61}{0.29} & \meanstd{89.38}{1.70} & \textbf{\meanstd{94.51}{0.31}} & \meanstd{86.99}{2.22} & \meanstd{93.45}{0.63} & \meanstd{84.21}{0.46} & \meanstd{90.95}{1.05} & \underline{\meanstd{94.31}{1.13}} \\
\cmidrule(lr){2-12}
& eTaP         & \meanstd{90.99}{0.00} & \textbf{\meanstd{93.46}{0.00}} & \underline{\meanstd{92.28}{1.16}} & \meanstd{91.17}{2.03} & \meanstd{80.86}{5.35} & \meanstd{83.39}{3.01} & \meanstd{89.75}{2.45} & \meanstd{34.23}{0.56} & \meanstd{87.44}{3.10} & \meanstd{91.26}{2.98} \\
& eTaR         & \meanstd{24.11}{0.00} & \meanstd{24.00}{0.00} & \meanstd{4.83}{0.70} & \meanstd{35.09}{5.22} & \meanstd{46.25}{1.82} & \meanstd{27.78}{6.02} & \underline{\meanstd{52.54}{3.44}} & \meanstd{10.77}{1.17} & \meanstd{40.27}{3.79} & \textbf{\meanstd{58.81}{4.26}} \\
& eTaF1        & \meanstd{38.12}{0.00} & \meanstd{38.19}{0.00} & \meanstd{9.17}{1.26} & \meanstd{50.53}{5.72} & \meanstd{58.78}{2.01} & \meanstd{41.46}{7.13} & \underline{\meanstd{66.21}{2.76}} & \meanstd{16.35}{1.44} & \meanstd{55.00}{3.49} & \textbf{\meanstd{71.45}{3.47}}\\
\midrule
\multirow[c]{10}{*}{WADI}
& P    & \meanstd{81.37}{0.00} & \meanstd{81.37}{0.00} & \meanstd{16.29}{20.17} & \meanstd{90.04}{0.91} & \meanstd{87.04}{2.97} & \meanstd{87.79}{2.28} & \underline{\meanstd{92.89}{0.25}} & \meanstd{55.13}{31.00} & \meanstd{92.62}{0.67} & \textbf{\meanstd{95.20}{0.44}} \\
& R       & \meanstd{13.14}{0.00} & \meanstd{13.14}{0.00} & \meanstd{0.86}{1.08} & \meanstd{26.82}{2.37} & \meanstd{21.08}{6.41} & \meanstd{22.49}{4.80} & \underline{\meanstd{38.23}{2.02}} & \meanstd{5.44}{3.13} & \meanstd{37.39}{5.47} & \textbf{\meanstd{58.05}{5.29}} \\
& F1           & \meanstd{22.63}{0.00} & \meanstd{22.63}{0.00} & \meanstd{1.64}{2.05} & \meanstd{41.29}{2.91} & \meanstd{33.64}{8.31} & \meanstd{35.64}{6.20} & \underline{\meanstd{54.14}{2.04}} & \meanstd{9.89}{5.69} & \meanstd{53.11}{5.56} & \textbf{\meanstd{72.02}{4.28}} \\
\cmidrule(lr){2-12}
& PA-P & \meanstd{84.29}{0.00} & \meanstd{84.29}{0.00} & \meanstd{27.35}{33.50} & \meanstd{96.00}{0.30} & \meanstd{95.46}{0.43} & \meanstd{93.60}{1.49} & \meanstd{95.74}{0.88} & \meanstd{67.35}{37.65} & \underline{\meanstd{96.02}{0.92}} & \textbf{\meanstd{96.84}{0.29}} \\
& PA-R    & \meanstd{16.15}{0.00} & \meanstd{16.15}{0.00} & \meanstd{2.69}{3.29} & \meanstd{70.95}{4.82} & \meanstd{62.77}{5.37} & \meanstd{46.18}{10.42} & \meanstd{68.02}{14.71} & \meanstd{12.92}{7.22} & \underline{\meanstd{74.42}{18.38}} & \textbf{\meanstd{89.85}{7.93}} \\
& PA-F1        & \meanstd{27.10}{0.00} & \meanstd{27.10}{0.00} & \meanstd{4.89}{6.00} & \meanstd{81.53}{3.31} & \meanstd{75.64}{4.00} & \meanstd{61.34}{9.96} & \meanstd{78.91}{9.78} & \meanstd{21.68}{12.12} & \underline{\meanstd{82.86}{12.70}} & \textbf{\meanstd{93.07}{4.58}} \\
\cmidrule(lr){2-12}
& eTaP         & \meanstd{46.50}{0.00} & \meanstd{47.05}{0.00} & \meanstd{10.61}{13.18} & \meanstd{82.27}{2.82} & \meanstd{77.55}{6.77} & \meanstd{84.37}{6.67} & \underline{\meanstd{87.07}{5.09}} & \meanstd{44.66}{24.99} & \meanstd{86.76}{7.45} & \textbf{\meanstd{89.34}{2.08}} \\
& eTaR         & \meanstd{19.61}{0.00} & \meanstd{19.61}{0.00} & \meanstd{1.89}{2.31} & \meanstd{46.99}{3.12} & \meanstd{44.60}{5.55} & \meanstd{33.96}{7.61} & \meanstd{55.34}{6.31} & \meanstd{12.26}{6.90} & \underline{\meanstd{55.49}{11.39}} & \textbf{\meanstd{71.61}{3.74}} \\
& eTaF1        & \meanstd{27.59}{0.00} & \meanstd{27.68}{0.00} & \meanstd{3.19}{3.91} & \meanstd{59.79}{3.21} & \meanstd{56.54}{5.73} & \meanstd{47.82}{7.36} & \underline{\meanstd{67.33}{3.33}} & \meanstd{19.23}{10.80} & \meanstd{66.76}{8.23} & \textbf{\meanstd{79.44}{2.37}} \\
\midrule
\multirow[c]{10}{*}{HAI}
& P     & \meanstd{50.37}{0.00} & \meanstd{60.59}{0.00} & \meanstd{35.42}{4.08} & \meanstd{84.42}{0.65} & \meanstd{78.85}{4.11} & \meanstd{79.06}{3.27} & \underline{\meanstd{86.00}{0.92}} & \meanstd{18.29}{3.51} & \meanstd{85.61}{1.52} & \textbf{\meanstd{87.44}{0.98}} \\
& R     & \meanstd{5.67}{0.00} & \meanstd{8.58}{0.00} & \meanstd{3.18}{0.58} & \meanstd{30.77}{1.54} & \meanstd{21.92}{5.45} & \meanstd{22.05}{4.22} & \underline{\meanstd{34.93}{2.57}} & \meanstd{1.27}{0.29} & \meanstd{34.07}{4.10} & \textbf{\meanstd{39.65}{3.35}} \\
& F1    & \meanstd{10.19}{0.00} & \meanstd{15.04}{0.00} & \meanstd{5.84}{1.02} & \meanstd{45.09}{1.75} & \meanstd{34.12}{7.06} & \meanstd{34.37}{5.50} & \underline{\meanstd{49.65}{2.77}} & \meanstd{2.37}{0.53} & \meanstd{48.66}{4.45} & \textbf{\meanstd{54.51}{3.41}} \\
\cmidrule(lr){2-12}
& PA-P  & \meanstd{69.96}{0.00} & \meanstd{86.46}{0.00} & \meanstd{58.78}{4.51} & \meanstd{90.14}{0.25} & \underline{\meanstd{93.11}{0.48}} & \meanstd{89.03}{0.96} & \meanstd{92.83}{0.20} & \meanstd{84.83}{6.87} & \meanstd{92.38}{0.56} & \textbf{\meanstd{93.55}{0.31}} \\
& PA-R  & \meanstd{13.00}{0.00} & \meanstd{35.67}{0.00} & \meanstd{8.32}{1.34} & \meanstd{51.82}{1.40} & \underline{\meanstd{76.67}{5.76}} & \meanstd{46.57}{4.70} & \meanstd{73.38}{2.20} & \meanstd{36.83}{15.24} & \meanstd{69.07}{5.60} & \textbf{\meanstd{82.33}{4.37}} \\
& PA-F1 & \meanstd{21.93}{0.00} & \meanstd{50.50}{0.00} & \meanstd{14.56}{2.22} & \meanstd{65.80}{1.21} & \underline{\meanstd{84.01}{3.71}} & \meanstd{61.06}{4.18} & \meanstd{81.96}{1.45} & \meanstd{50.27}{16.87} & \meanstd{78.96}{3.84} & \textbf{\meanstd{87.54}{2.56}} \\
\cmidrule(lr){2-12}
& eTaP  & \meanstd{47.19}{0.00} & \meanstd{48.47}{0.00} & \meanstd{62.55}{2.97} & \meanstd{78.56}{1.59} & \meanstd{72.08}{3.03} & \meanstd{65.62}{9.82} & \underline{\meanstd{80.17}{5.35}} & \meanstd{18.29}{3.51} & \meanstd{77.79}{2.17} & \textbf{\meanstd{84.00}{3.53}} \\
& eTaR  & \meanstd{11.63}{0.00} & \meanstd{16.28}{0.00} & \meanstd{7.22}{0.50} & \meanstd{21.30}{1.87} & \underline{\meanstd{46.60}{5.47}} & \meanstd{22.38}{5.79} & \meanstd{43.69}{2.79} & \meanstd{11.23}{2.41} & \meanstd{41.17}{2.19} & \textbf{\meanstd{51.93}{1.97}} \\
& eTaF1 & \meanstd{18.66}{0.00} & \meanstd{24.37}{0.00} & \meanstd{12.94}{0.84} & \meanstd{33.48}{2.41} & \meanstd{56.48}{4.58} & \meanstd{33.22}{7.40} & \underline{\meanstd{56.51}{3.13}} & \meanstd{13.90}{2.84} & \meanstd{53.84}{2.39} & \textbf{\meanstd{64.15}{2.04}} \\
\midrule
\multicolumn{2}{l}{Avg. rank $\downarrow$} 
& 7.67 & 7.11 & 9.78 & 4.33 & 4.11 & 5.67 & \underline{2.44} & 9.11 & 3.33 & \textbf{1.11} \\
\bottomrule
\end{tabularx}
\end{table*}

\subsection{EQ1. Overall Performance}\label{sec:eq1}
Table~\ref{tbl:full_results} shows the anomaly detection performance of all compared methods on the three datasets.
Overall, {\method} achieves the best average rank of 1.11 across F1, PA-F1, and eTaF1, outperforming all nine baselines.
Specifically, {\method} achieves the best performance on 24 out of 27 dataset--metric pairs, demonstrating consistently superior performance across diverse datasets and evaluation metrics.
The improvements over the second-best method, FuSAGNet, in F1 and eTaF1 are statistically significant on all three datasets (i.e., the $p$-values are below 0.05).
Notably, on WADI, {\method} outperforms the best-performing baseline for each metric by 17.88, 10.21, and 12.11 percentage points in F1, PA-F1, and eTaF1, respectively.
These results demonstrate the effectiveness of {\method} across different datasets and evaluation settings.

\vspace{1mm}
\noindent\textbf{Precision--recall characteristics.}
A closer examination of precision and recall reveals that {\method} consistently improves recall while maintaining high precision.
On WADI, {\method} achieves a recall of 58.05, exceeding the best baseline result of 38.23 by 19.82 percentage points, while also improving precision from 92.89 to 95.20.
Similar trends are observed on SWaT and HAI, where {\method} achieves the highest precision and recall among all compared methods.
These results indicate that {\method} detects more anomalous instances without sacrificing precision.
This balance between precision and recall contributes to its superior F1 performance, particularly on WADI, where existing methods exhibit relatively low recall.

\vspace{1mm}
\noindent\textbf{Performance across evaluation metrics.}
{\method} also demonstrates consistent effectiveness under different evaluation protocols.
While conventional F1 evaluates point-wise detection accuracy, PA-F1 relaxes localization errors by adjusting predictions within anomalous segments.
As a result, PA-F1 may not fully reflect the accuracy of individual anomaly predictions.
In contrast, eTaF1 further considers temporal characteristics of anomaly detection.
{\method} achieves the highest F1 and eTaF1 on all three datasets, indicating that its superior performance is not limited to point-adjusted evaluation.
Although TSMixer achieves a slightly higher PA-F1 on SWaT, {\method} outperforms it in both F1 and eTaF1.
These results further demonstrate the superior anomaly detection performance of {\method} across multiple complementary evaluation metrics.

\begin{figure*}[tp]
    \centering
    \setlength{\tabcolsep}{1mm}
    \begin{tabular}{@{}ccc@{}}
        \multicolumn{3}{c}{
            \includegraphics[width=0.5\linewidth]
            {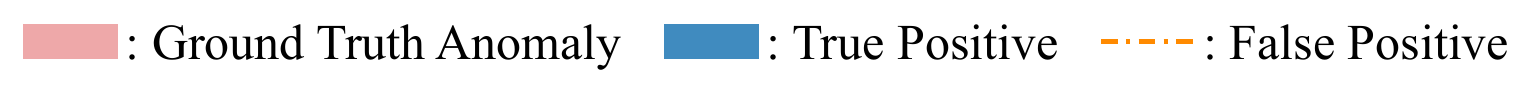}
        } \\
        \includegraphics[width=0.33\linewidth]
        {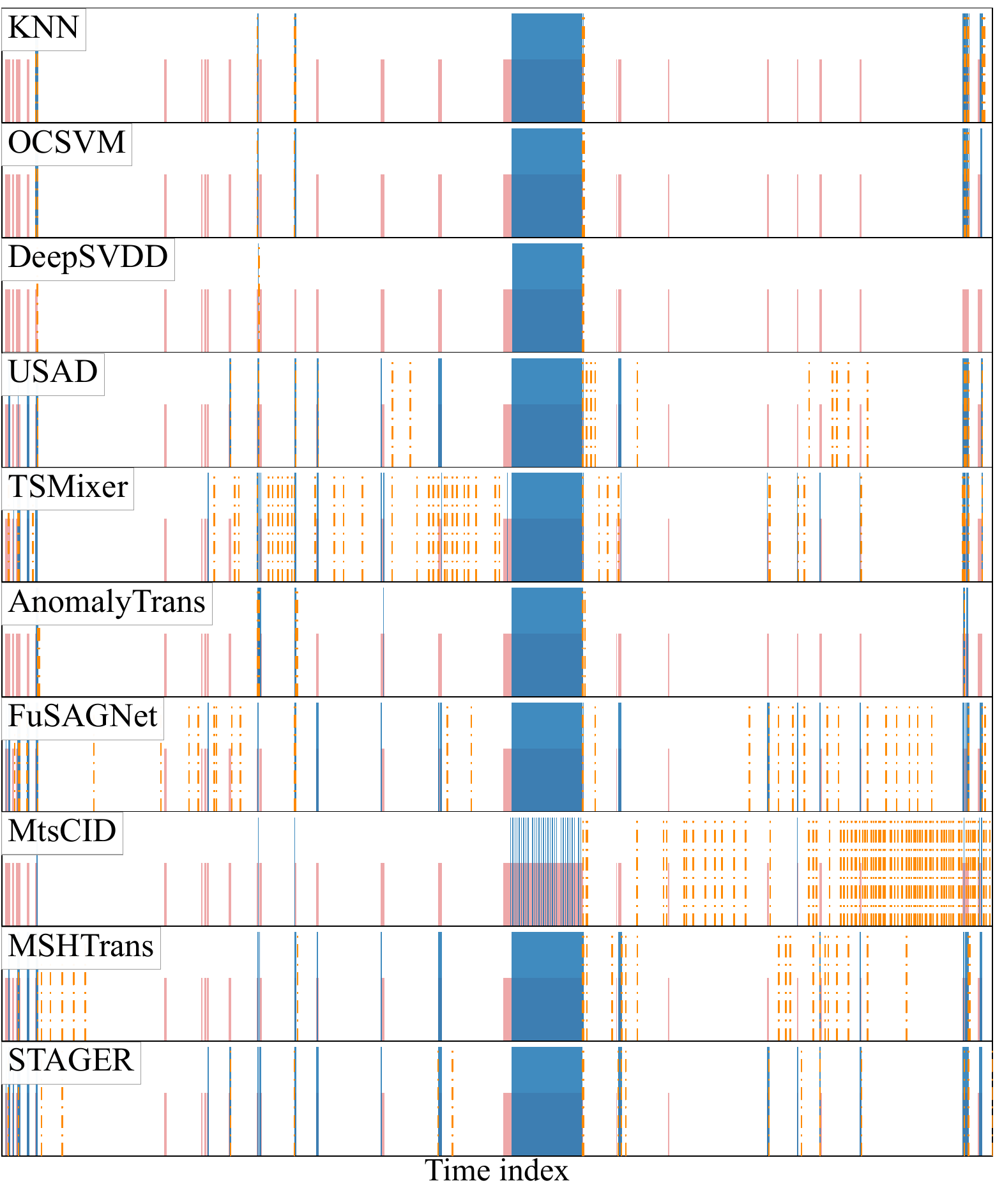} &
        \includegraphics[width=0.33\linewidth]
        {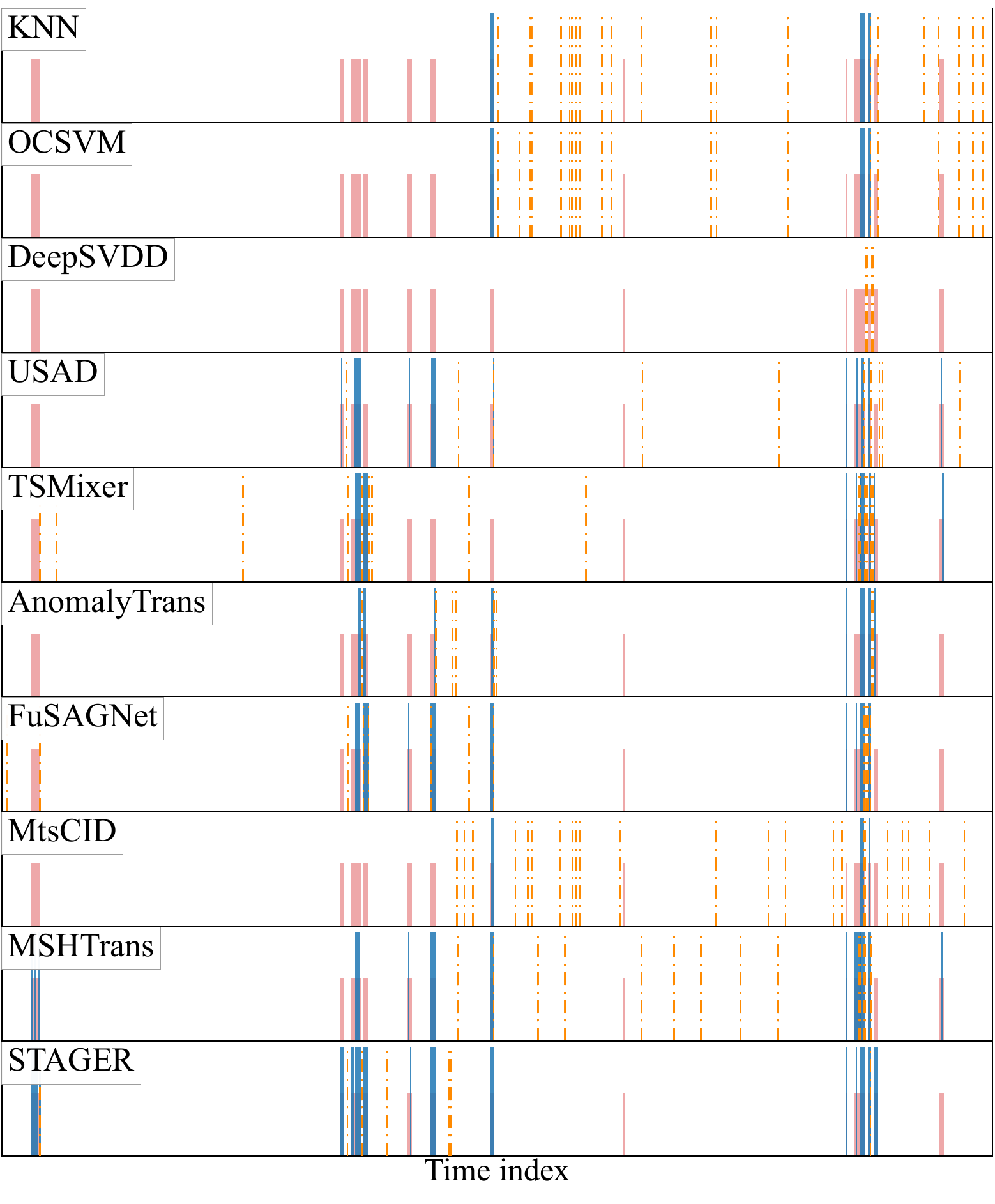} &
        \includegraphics[width=0.33\linewidth]
        {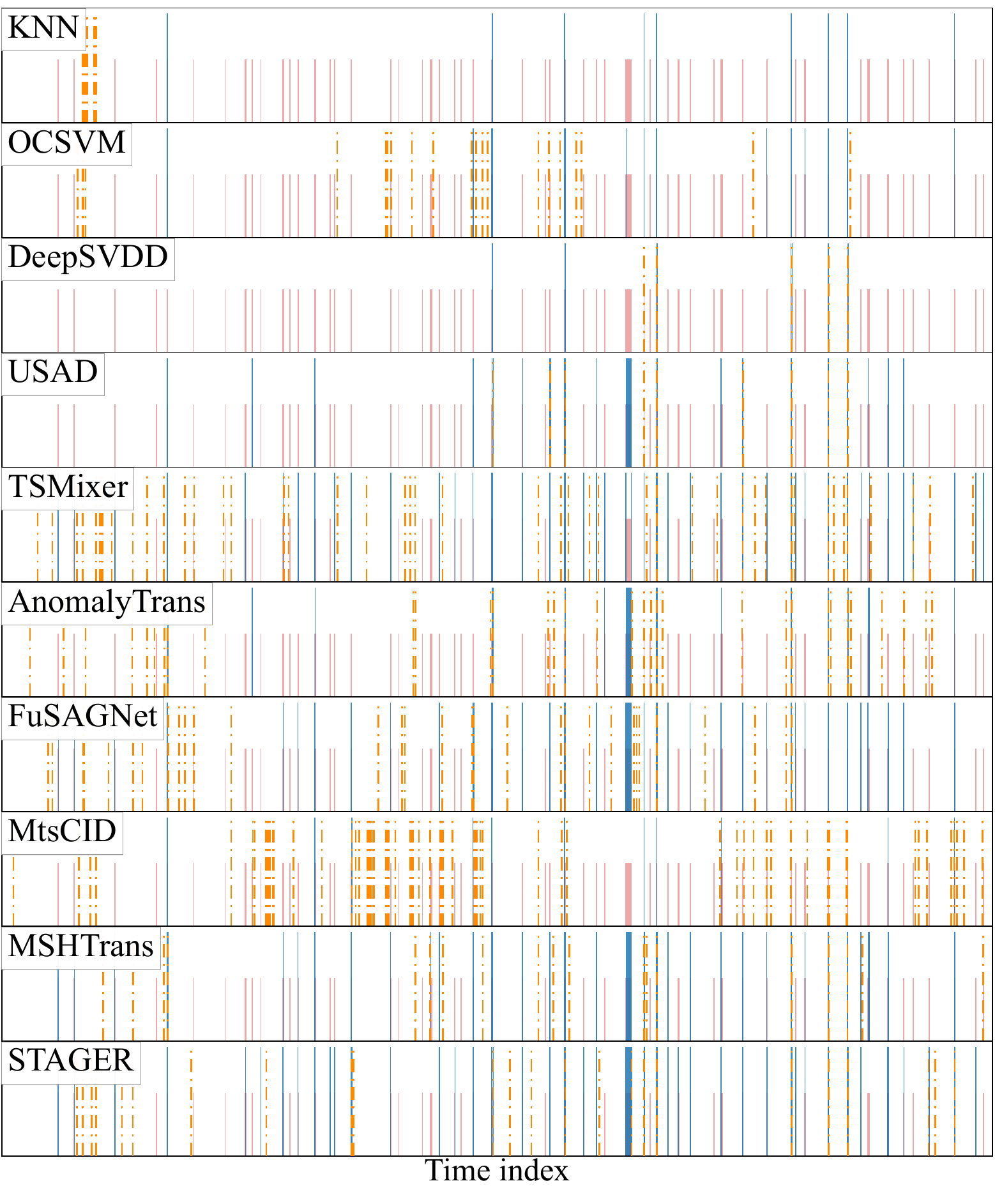} \\
        {\small (a) SWaT} &
        {\small (b) WADI} &
        {\small (c) HAI}
    \end{tabular}
    \caption{Qualitative comparison of anomaly detection results on the SWaT, WADI, and HAI test sets.}
    \label{fig:pred_qualitative_comparison}
    \vspace{-4mm}
\end{figure*}

\begin{table*}[t]
\caption{Ablation study of {\method} on SWaT, WADI, and HAI. We evaluate the contributions of (1) the scale-robust temporal DAG encoder and (2) the global-aware relational encoder through component removal and alternative implementations. The \textbf{bold} and \underline{underline} indicate the full model and the best-performing ablation variant within each encoder group, respectively.}

\label{tbl:ablation}
\centering
\footnotesize
\setlength{\tabcolsep}{2pt}
\def\arraystretch{1.05} 
\begin{tabularx}{\textwidth}{@{}clYYYYYYYYY@{}}
\toprule
& \multirow{2}{*}{Method}
& \multicolumn{3}{c}{SWaT}
& \multicolumn{3}{c}{WADI}
& \multicolumn{3}{c}{HAI} \\
\cmidrule(lr){3-5}
\cmidrule(lr){6-8}
\cmidrule(lr){9-11}
& & F1 & PA-F1 & eTaF1
& F1 & PA-F1 & eTaF1
& F1 & PA-F1 & eTaF1 \\
\midrule

& \method{} (FiLM+GAT)
& \textbf{\meanstd{87.29}{1.01}} & \textbf{\meanstd{94.31}{1.13}} & \textbf{\meanstd{71.45}{3.47}}
& \textbf{\meanstd{72.02}{4.28}} & \textbf{\meanstd{93.07}{4.58}} & \textbf{\meanstd{79.44}{2.37}}
& \textbf{\meanstd{54.51}{3.41}} & \textbf{\meanstd{87.54}{2.56}} & \textbf{\meanstd{64.15}{2.04}} \\

\midrule

\multirow{2}{*}{\textbf{(1)}}
& \method{} w/o MS
& \underline{\meanstd{86.49}{0.98}} & \underline{\meanstd{93.65}{1.02}} & \underline{\meanstd{68.64}{3.28}}
& \underline{\meanstd{59.68}{4.09}} & \underline{\meanstd{91.44}{4.67}} & \underline{\meanstd{75.29}{2.07}}
& \meanstd{41.32}{7.07} & \meanstd{81.59}{1.92} & \meanstd{52.91}{2.31} \\

& \method{} w/o DAG
& \meanstd{86.26}{2.93} & \meanstd{93.11}{3.80} & \meanstd{64.88}{11.08}
& \meanstd{56.87}{3.20} & \meanstd{84.01}{8.47} & \meanstd{70.94}{3.09}
& \underline{\meanstd{53.51}{7.92}} & \underline{\meanstd{85.48}{2.54}} & \underline{\meanstd{62.99}{2.58}} \\

\midrule

\multirow{4}{*}{\textbf{(2)}}
& \method{} w/o Global
& \meanstd{81.53}{0.65} & \meanstd{88.50}{1.01} & \meanstd{55.10}{1.62}
& \meanstd{35.23}{3.41} & \meanstd{42.63}{7.10} & \meanstd{45.74}{4.85}
& \underline{\meanstd{52.87}{6.07}} & \meanstd{84.84}{3.27} & \underline{\meanstd{60.61}{2.97}} \\

& \method{} (MLP\_Global)
& \underline{\meanstd{86.12}{2.32}} & \meanstd{92.85}{2.89} & \meanstd{66.44}{9.35}
& \meanstd{60.16}{3.05} & \underline{\meanstd{90.77}{4.30}} & \meanstd{73.85}{3.47}
& \meanstd{51.29}{6.64} & \underline{\meanstd{84.97}{5.94}} & \meanstd{60.32}{4.98} \\

& \method{} (GCN)
& \meanstd{84.49}{0.40} & \meanstd{93.37}{1.34} & \meanstd{63.37}{5.93}
& \meanstd{54.87}{7.77} & \meanstd{84.87}{18.19} & \meanstd{71.03}{11.03}
& \meanstd{47.19}{1.76} & \meanstd{76.11}{0.92} & \meanstd{49.51}{1.20} \\

& \method{} (MLP\_Rel)
& \meanstd{85.63}{0.87} & \underline{\meanstd{94.28}{0.81}} & \underline{\meanstd{67.62}{2.04}}
& \underline{\meanstd{67.10}{3.55}} & \meanstd{88.70}{2.78} & \underline{\meanstd{74.23}{2.51}}
& \meanstd{48.96}{1.53} & \meanstd{79.91}{5.14} & \meanstd{57.66}{3.32} \\

\bottomrule
\end{tabularx}
\end{table*}

\vspace{1mm}
\noindent\textbf{Qualitative comparison.}
To better understand the prediction behavior of {\method} and competing methods, we qualitatively compare their anomaly detection results.
Figure~\ref{fig:pred_qualitative_comparison} shows the predictions of all methods on the three test sets.
On SWaT, although several methods successfully detect the prominent anomalous segment, methods such as TSMixer and MtsCID produce more fragmented false alarms across normal periods.
In contrast, {\method} exhibits a more concentrated detection pattern with relatively fewer false positives.
On WADI, several baselines produce numerous false alarms or fail to detect substantial portions of anomalous segments, whereas {\method} identifies anomalous instances with comparatively fewer false alarms.
A similar tendency is observed on HAI, where {\method} detects anomalous instances distributed across the test period while maintaining a relatively balanced prediction pattern.
These qualitative observations are consistent with the quantitative results, particularly the improvements in recall and F1.

Overall, the quantitative and qualitative results demonstrate that {\method} achieves consistently superior anomaly detection performance across different datasets and evaluation metrics.
In particular, {\method} improves the detection of anomalous instances while maintaining high precision, resulting in more accurate and reliable anomaly detection.

\begin{table*}[t]
\caption{Effectiveness of the proposed auxiliary objectives. Jointly optimizing both objectives achieves the best scores across all evaluation metrics on all three datasets. The \textbf{bold} and \underline{underline} indicate the best and second-best results, respectively.}
\label{tbl:auxiliary}
\centering
\footnotesize
\setlength{\tabcolsep}{1pt}
\def\arraystretch{1.2} 
\begin{tabularx}{\textwidth}{@{}lccYYYYYYYYY@{}}
\toprule
\multirow{2}{*}{Method}
& \multicolumn{2}{c}{Aux.}
& \multicolumn{3}{c}{SWaT}
& \multicolumn{3}{c}{WADI}
& \multicolumn{3}{c}{HAI} \\
\cmidrule(lr){2-3}
\cmidrule(lr){4-6}
\cmidrule(lr){7-9}
\cmidrule(lr){10-12}
& Cls. & Reg.
& F1 & PA-F1 & eTaF1
& F1 & PA-F1 & eTaF1
& F1 & PA-F1 & eTaF1 \\
\midrule

\method{}
& \checkmark & \checkmark
& \textbf{\meanstd{87.29}{1.01}} & \textbf{\meanstd{94.31}{1.13}} & \textbf{\meanstd{71.45}{3.47}}
& \textbf{\meanstd{72.02}{4.28}} & \textbf{\meanstd{93.07}{4.58}} & \textbf{\meanstd{79.44}{2.37}}
& \textbf{\meanstd{54.51}{3.41}} & \textbf{\meanstd{87.54}{2.56}} & \textbf{\meanstd{64.15}{2.04}} \\

\midrule

\method{} w/o Cls.
& - & \checkmark
& \meanstd{86.29}{2.53} & \underline{\meanstd{94.21}{2.31}} & \underline{\meanstd{67.71}{10.84}}
& \underline{\meanstd{64.51}{6.65}} & \meanstd{89.86}{8.36} & \underline{\meanstd{76.06}{3.14}}
& \meanstd{46.85}{7.06} & \underline{\meanstd{86.74}{0.99}} & \underline{\meanstd{63.14}{1.84}} \\

\method{} w/o Reg.
& \checkmark & -
& \meanstd{85.84}{2.82} & \meanstd{92.86}{2.68} & \meanstd{61.94}{10.41}
& \meanstd{58.52}{0.75} & \meanstd{89.26}{2.77} & \meanstd{72.35}{1.68}
& \meanstd{51.27}{5.98} & \meanstd{84.76}{5.34} & \meanstd{61.49}{6.47} \\

\method{} w/o Both
& - & -
& \underline{\meanstd{86.36}{1.69}} & \meanstd{93.16}{1.72} & \meanstd{65.60}{7.30}
& \meanstd{58.80}{4.98} & \underline{\meanstd{92.99}{1.88}} & \meanstd{73.64}{2.70}
& \underline{\meanstd{52.05}{3.43}} & \meanstd{86.20}{3.26} & \meanstd{62.99}{2.09} \\

\bottomrule
\end{tabularx}
\end{table*}

\subsection{EQ2. Ablation Study}\label{sec:eq2}
We evaluate the effectiveness of the two proposed encoders through six ablation variants on SWaT, WADI, and HAI, as summarized in Table~\ref{tbl:ablation}. 
For the scale-robust temporal DAG encoder, 
{\method} w/o MS removes multi-scale representations and uses only raw observations, while {\method} w/o DAG replaces the temporal DAG with a fully connected graph while retaining multi-scale representations. 
For the global-aware relational encoder, {\method} w/o Global removes the series-level global embeddings and their integration, whereas {\method} (MLP\_Global) replaces FiLM with MLP-based fusion. 
To evaluate inter-series relational modeling, {\method} (GCN) and {\method} (MLP\_Rel) replace the relational GAT with a GCN and an MLP-based transformation, respectively. 
Overall, the full {\method} achieves the best scores across all evaluation metrics on all three datasets, demonstrating the effectiveness of combining multi-scale temporal learning with global-aware relational modeling.

\vspace{1mm}
\noindent
\textbf{(1) Scale-robust temporal DAG encoder}. We evaluate the proposed temporal encoder by removing either the per-series multi-scale representation (STAGER w/o MS) or the temporal DAG (STAGER w/o DAG). 
Removing the multi-scale representation reduces F1 and eTaF1 across all three datasets, indicating the benefit of capturing temporal patterns at multiple scales rather than relying solely on raw observations. 
Replacing the temporal DAG with a fully connected graph also degrades F1 and eTaF1 across all datasets, suggesting that explicitly preserving chronological dependencies is more effective than allowing unrestricted information propagation across time steps. 
However, the relative contributions of the two components vary across datasets.
The temporal DAG contributes more substantially on SWaT and WADI, whereas multi-scale representation learning has a greater impact on HAI. 
These findings suggest that the two components provide complementary benefits for learning temporal representations under different time-series characteristics.

\vspace{1mm}
\noindent
\textbf{(2) Global-aware relational encoder.} We evaluate this encoder from two perspectives: global context integration and inter-series relational modeling. 
First, we examine the contribution of series-level global embeddings and the effectiveness of their integration into temporal representations.
Removing global embeddings substantially reduces detection performance, particularly on SWaT and WADI, where eTaF1 decreases from 71.45 to 55.10 and from 79.44 to 45.74, respectively. 
This indicates that series-level global characteristics provide complementary information beyond window-specific temporal patterns. 
Furthermore, FiLM achieves higher eTaF1 than MLP-based fusion across all three datasets, suggesting that feature-wise modulation is effective for integrating global context into temporal representations.

Next, we examine inter-series relational modeling by comparing GAT-based aggregation with GCN-based and MLP-based aggregation.
The GAT-based encoder achieves higher eTaF1 than both alternatives across all three datasets, demonstrating the benefit of attention-based relational modeling. 
Notably, the MLP variant outperforms GCN in eTaF1 across all datasets, suggesting that uniform graph propagation does not necessarily improve relational representations. 
In contrast, GAT adaptively weights information from different series, allowing the encoder to account for varying strengths of inter-series dependencies.
Together, these results highlight the complementary roles of global context integration and adaptive relational modeling in learning representations that capture both persistent series-level characteristics and window-specific inter-series dependencies.

\begin{figure*}[pos=tbp]
    \centering
    \begin{tabular}{@{}c@{}}
        \includegraphics[width=0.6\linewidth]{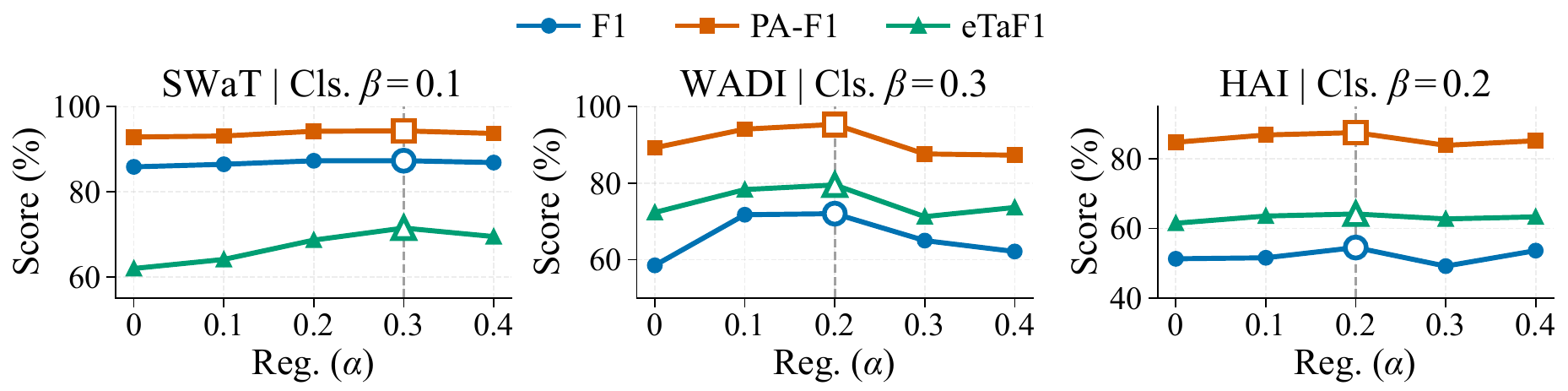} 
        \\
        \includegraphics[width=0.6\linewidth]{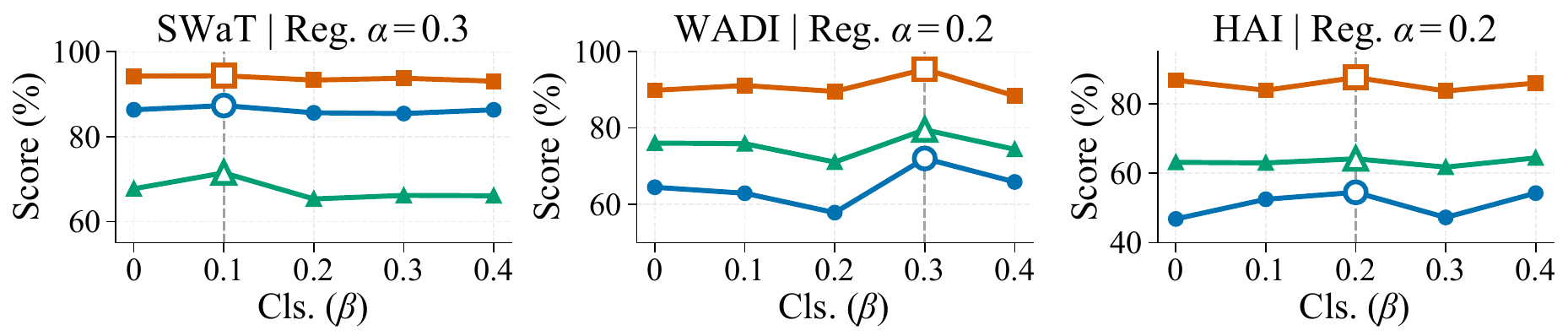}
    \end{tabular}
    \caption{Sensitivity analysis of {\method} to the auxiliary loss weights: regularization $\alpha$ (top) and classification $\beta$ (bottom).}
    \label{fig:hyperparameter-sensitivity}
    \vspace{-3mm}
\end{figure*}


\begin{figure*}[pos=tbp]
    \centering
    \begin{tabular}{@{}c@{}}
        \includegraphics[width=0.6\linewidth]
        {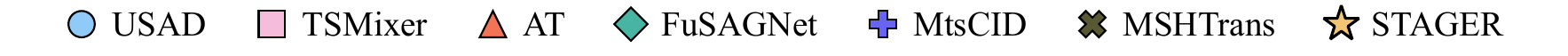} \\
        \includegraphics[width=0.6\linewidth]
        {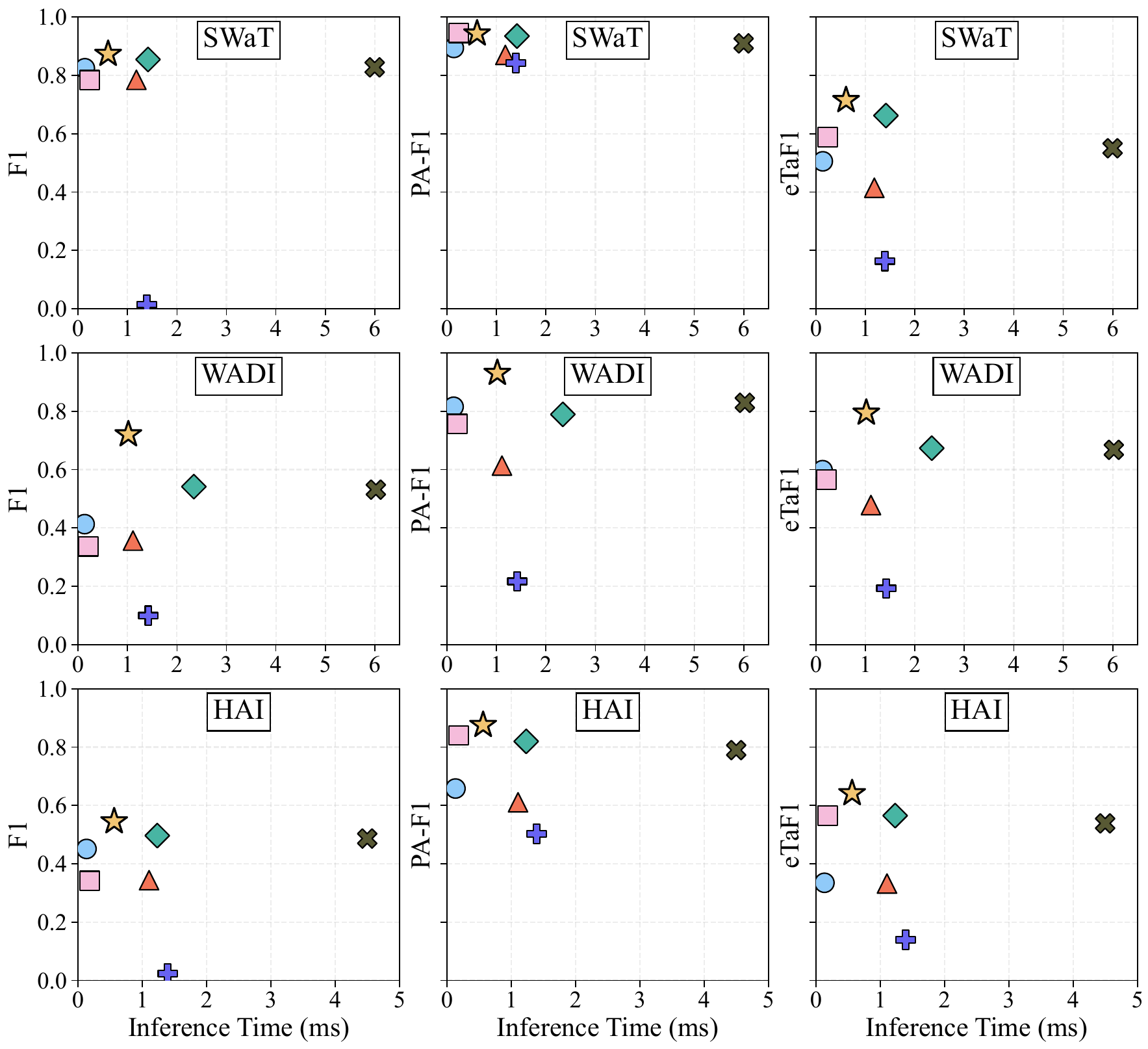}
    \end{tabular}
    \vspace{-2mm}
    \caption{Trade-off between detection performance and inference time of different MTSAD methods on the SWaT, WADI, and HAI datasets. Methods closer to the upper-left exhibit a better balance between detection performance and inference time.}
    \label{fig:inference_time}
    \vspace{-3mm}
\end{figure*}

\subsection{EQ3. Effectiveness of Auxiliary Objectives}\label{sec:eq3}
We evaluate the effectiveness of the two auxiliary objectives for learning series-level global embeddings: (i) series classification and (ii) distance-based regularization. 
Table~\ref{tbl:auxiliary} presents the results of three ablation variants on SWaT, WADI, and HAI. 
Specifically, {\method} w/o Cls. removes the series classification objective,
{\method} w/o Reg. removes the distance-based regularization objective, 
and {\method} w/o Both removes both objectives. 
Overall, the full {\method} achieves the best F1, PA-F1, and eTaF1 scores across all three datasets, demonstrating the benefit of jointly optimizing the two auxiliary objectives.

Removing the classification objective consistently degrades detection performance, with eTaF1 decreasing from 71.45 to 67.71 on SWaT, from 79.44 to 76.06 on WADI, and from 64.15 to 63.14 on HAI. 
These results suggest that encouraging distinguishability among series-level global embeddings helps capture series-specific characteristics that complement window-specific temporal representations.
Removing distance-based regularization leads to larger performance degradation, particularly on SWaT and WADI. 
Notably, {\method} w/o Reg. achieves lower eTaF1 than  w/o Both across all three datasets. 
This finding suggests that classification alone may encourage series distinguishability without adequately preserving meaningful relationships among global embeddings, potentially limiting their usefulness for anomaly detection.

Together, these results highlight the complementary roles of the two objectives: series classification encourages discriminative global embeddings, while distance-based regularization helps maintain their structural relationships. Their joint optimization enables more informative series-level global representations, contributing to improved anomaly detection performance.

\vspace{1mm}
\noindent
\textbf{Sensitivity analysis.}
We conduct a sensitivity analysis of the auxiliary loss weights, $\alpha$ and $\beta$, by varying one weight while fixing the other at its selected value. 
Figure~\ref{fig:hyperparameter-sensitivity} presents the results across the three evaluation metrics.
Overall, the sensitivity to the auxiliary loss weights varies by dataset. 
On SWaT, increasing $\alpha$ generally improves detection performance up to $\alpha=0.3$, while varying $\beta$ has a relatively limited effect. 
On WADI, performance is relatively more sensitive to both weights, with the highest scores across all three metrics observed at $\alpha=0.2$ and $\beta=0.3$.
In contrast, HAI exhibits relatively stable performance across different weight settings, with moderate improvements around $\alpha=0.2$ and $\beta=0.2$.
Based on these results, we set $(\alpha,\beta)$ to $(0.3, 0.1)$, $(0.2, 0.3)$, and $(0.2, 0.2)$ for SWaT, WADI, and HAI, respectively.

\subsection{EQ4. Detection Efficiency}\label{sec:eq4}
Finally, we compare the trade-off between detection performance and inference time of {\method} and six representative neural baselines by plotting F1, PA-F1, and eTaF1 against per-window inference time.
Figure~\ref{fig:inference_time} shows the results, where methods located toward the upper-left corner achieve higher detection performance with lower inference time.
Overall, {\method} achieves a favorable balance between detection performance and inference efficiency across all three datasets, attaining the highest F1 and eTaF1 scores while maintaining low inference time.
Although USAD and TSMixer are slightly faster, their detection performance is generally lower than that of {\method}.

Other methods exhibit different trade-offs. 
AnomalyTransformer provides relatively fast inference but lower detection performance, whereas MSHTrans requires substantially longer inference time without achieving comparable detection performance.
Among the competing methods, FuSAGNet achieves competitive detection performance but requires longer inference time than {\method}, 
while its performance advantage over faster baselines is not consistent across datasets and evaluation metrics.

These results demonstrate that {\method} achieves high detection performance while maintaining practical inference efficiency, 
making it a promising choice for time-sensitive anomaly detection applications.

\section{Conclusion and Future Work}\label{sec:con}
In this paper, we identify two key challenges in multivariate time series anomaly detection: (C1) learning scale-robust temporal representations and (C2) capturing series-level global characteristics beyond local time windows.
To address these challenges, we propose {\method}, which integrates (1) a scale-robust temporal DAG encoder and (2) a global-aware relational encoder 
to capture multi-scale temporal patterns, series-level global characteristics, and inter-series dependencies.
We further introduce two auxiliary objectives to facilitate the learning of informative series-level global embeddings.
Extensive experiments on three real-world benchmark datasets demonstrate that {\method} achieves the best overall detection performance among the evaluated MTSAD methods in terms of average rank across multiple evaluation metrics.
Furthermore, ablation and sensitivity analyses validate the contributions of the proposed components and auxiliary objectives, while efficiency evaluation demonstrates that {\method} achieves a favorable trade-off between detection performance and inference time.

Although the learned global profiles provide consistent series-level information across input windows, they remain fixed during inference and may not fully capture evolving series characteristics under distribution shifts.
In addition, the complete inter-series graph may incur increasing computational overhead as the number of series grows, particularly in high-dimensional multivariate time series.
In future work, we plan to investigate adaptive global profile updating and sparse, dynamic inter-series graph construction to better accommodate evolving series characteristics and inter-series dependencies while improving scalability to larger systems.


\printcredits

\section*{Acknowledgements}
This work was supported by the National Research Foundation of Korea (NRF) grant funded by the Korea government (MSIT) (RS-2024-00459301), the affiliated institute of ETRI[2026-070], and the Chung-Ang University Graduate Research Scholarship in 2025. 

\section*{Declaration of Generative AI and AI-assisted Technologies in the Manuscript Preparation Process}
During the preparation of this work, the authors used ChatGPT (OpenAI) to assist with language editing and improving the clarity and structure of the manuscript.
After using this tool, the authors reviewed and edited the content as needed and take full responsibility for the content of the published article.

\bibliographystyle{cas-model2-names}

\bibliography{KBS-STAGER}



\end{document}